\documentclass[twoside]{article}

\usepackage[accepted]{aistats2022}
\usepackage[round]{natbib}

\usepackage{mathtools}
\usepackage{amsmath,amscd,amssymb}
\usepackage{bbold}
\usepackage{bm}
\usepackage[linesnumbered,ruled,vlined]{algorithm2e}
\usepackage[many]{tcolorbox}
\usepackage{enumitem}
\usepackage{bbding}
\usepackage{floatrow}

\usepackage{url}            
\usepackage{booktabs}       
\usepackage{nicefrac}       
\usepackage{microtype}      
\usepackage{grffile}

\usepackage{graphicx}
\usepackage{xcolor,colortbl}
\usepackage{tabu}
\usepackage{hyperref}

\newcommand{\by}{\mathbf{y}}

\newcommand{\bomega}{\bm{\omega}}
\newcommand{\bOmega}{\bm{\Omega}}
\newcommand{\bff}{\mathbf{f}}
\newcommand{\bx}{\mathbf{x}}
\newcommand{\bz}{\mathbf{z}}
\newcommand{\bX}{\mathbf{X}}
\newcommand{\EE}{\mathbb{E}}
\newcommand{\DD}{\mathcal{D}}
\newcommand{\OO}{\mathcal{O}}
\newcommand{\II}{\mathcal{I}}
\newcommand{\RR}{\mathbb{R}}
\newcommand{\bK}{\bm{K}}

\newcommand{\olya}{\`olya}
\newcommand{\LLMLL}{\mathcal{L}_{\rm MLL}}
\newcommand{\LL}{\mathcal{L}}
\newcommand{\LLCV}{\mathcal{L}_{\rm CV}}
\newcommand{\LLL}{\mathcal{L}_{\rm LOO}}
\newcommand{\LLLst}{\widehat{\LL}^k_{\rm LOO}}
\newcommand{\NN}{\mathcal{N}}
\newcommand{\sigmaobs}{\sigma_{\rm obs}}
\newcommand{\thalf}{\tfrac{1}{2}}

\begin{document}

%

%

\twocolumn[

\aistatstitle{Scalable Cross Validation Losses for Gaussian Process Models}

\aistatsauthor{ Martin Jankowiak \And  Geoff Pleiss }

\aistatsaddress{ Broad Institute \And  Columbia University } ]

\begin{abstract}
We introduce a simple and scalable method for training Gaussian process (GP) models that exploits cross-validation and nearest neighbor truncation.
To accommodate binary and multi-class classification we leverage P\olya-Gamma auxiliary variables and variational inference.
In an extensive empirical comparison with a number of alternative methods for scalable GP regression and classification,
we find that our method offers fast training and excellent predictive performance.
We argue that the good predictive performance can be traced to the non-parametric nature of the resulting predictive distributions as
well as to the cross-validation loss, which provides robustness against model mis-specification.
\end{abstract}

\section{Introduction}
\label{sec:intro}

As machine learning becomes more widely used, it is increasingly being deployed in applications 
where autonomous decisions are guided by predictive models. 
For example, supply forecasts can determine the prices charged by retailers,  
expected demand for transportation can be used to optimize bus schedules, and 
data-driven algorithms can guide load balancing in critical electrical subsystems.
For these and many other applications of machine learning, it is essential that models are well-calibrated, thus
enabling downstream decisions to factor in uncertainty and risk.

Gaussian processes are a general-purpose modeling component that offer excellent 
uncertainty quantification in a variety of predictive tasks, including regression, classification, and beyond 
\citep{rasmussen2003gaussian}. 
Despite their many attractive features, wider use of Gaussian process (GP) models is hindered by 
computational requirements that can be prohibitive.
For example, classic methods for training GP regressors using the marginal log likelihood (MLL) 
scale cubically with the number
of data points. This bottleneck has motivated extensive research into approximate GP inference schemes
with more favorable computational properties \citep{liu2020gaussian}. 

A particularly popular and fruitful approach has centered on inducing point methods and variational inference 
\citep{snelson2006sparse,titsias2009variational,hensman2013gaussian}. These methods
rely on the ELBO, which is a lower bound to the MLL, for training and trade cubic complexity in the size
of the dataset $N$ for cubic complexity in the number of inducing points $M$. 
Since $M$ is a hyperparameter controlled by the user, inducing point methods can lead to substantial speed-ups.
One disadavantage of inducing point methods---as we will see explicitly in our empirical evaluation---is that
some datasets may require prohibitively large values of $M$ to ensure good model fit.

Instead of targeting the MLL directly or via a lower bound, we investigate the suitability of training objectives
based on \emph{cross-validation} (CV). In this work we argue that CV is attractive in the context
of GP models for three reasons in particular: 
i) it can provide robustness against model mis-specification; 
ii) it opens the door to simple approximation schemes based on nearest neighbor truncation;
and iii) nearest neighbor truncation enables non-parametric predictive distributions that avoid
some of the disadvantages of inducing point methods.
To put it differently, although nearest neighbors are a natural starting point for scalable GP methods,
their combination with MLL-based objectives is typically made awkward by the need to specify an ordering of the data
\citep{vecchia1988estimation, datta2016hierarchical}.
In contrast, CV-based objectives are free of any such requirement, resulting in an attractive synergy between
cross-validation and nearest neighbor approximations.
To exploit this synergy we make the following contributions:
\begin{enumerate}[topsep=0pt,itemsep=-0.5ex,partopsep=1ex,parsep=1ex]
\item We introduce the $k$-nearest-neighbor leave-one-out ({\bf LOO-}$\mathbf k$) objective, which can be used 
        to train GP models on large datasets.
\item We introduce a P\olya-Gamma auxiliary variable construction that extends this approach to binary and multi-class classification.
\item We perform an extensive empirical comparison with a number of alternative methods for scalable GP regression and classification
    and demonstrate the excellent predictive performance of our approach.
\end{enumerate}
Before we describe our approach in Sec.~\ref{sec:cv} we first review some basic background on Gaussian processes, which also gives
us the opportunity to establish some of the notation we will use throughout.

\section{Background on Gaussian processes}
\label{sec:bg}

\label{sec:gp}

A GP on the input space $\DD \subset \RR^D $
is specified\footnote{Unless otherwise noted we assume that the prior mean is uniformly zero.}
 by a covariance function or kernel $K:\DD\times \DD \to \mathbb{R}$ \citep{rasmussen2003gaussian}.
 A common choice is the RBF or squared exponential kernel, which is given by
\begin{equation}
K(\bx, \bz) = \sigma_K^2 \exp \{ -\tfrac{1}{2} \Sigma_i (x_i - z_i)^2 / \rho_i^2 \}
\end{equation}
where $\{\rho_i \}$ are length scales and $\sigma_K$ is the kernel scale.
For scalar regression $f: \DD \to \mathbb{R}$ the joint density of a GP regressor takes the form
\begin{equation}
p(\by, \bff | \bX) = \NN(\by|\bff, \sigmaobs^2 \mathbb{1}_N) \NN(\bff | \bm{0}, \bK_{N,N})
\end{equation}
where $\by$ are the real-valued targets, $\bff$ are the latent function values,
$\bX = \{ \bx_i \}_{i=1}^N$ are the $N$ inputs with $\bx_i \in \DD$,
$\sigmaobs^2$ is the variance of the Normal likelihood,
and $\bK_{N,N}$ is the $N \times N$ kernel matrix.
The marginal log likelihood (MLL) $\log p(\by|\bX ) = \log {\textstyle \int} d \bff \; p(\by, \bff | \bX)$  of the observed data can be computed in closed form:
\begin{align}
    \label{eqn:mll}
    \log p(\by|\bX ) = \log \NN(\by, \bK_{N,N} + \sigmaobs^2 \mathbb{1}_N)
\end{align}
Computing $\log p(\by|\bX )$ has cost $\OO(N^3)$, which has motivated the great variety of approximate methods
for scalable training of GP models \citep{liu2020gaussian}.
The posterior predictive distribution $p(y_* | \by, \bX, \bx_*)$ of the GP 
at a test point $\bx_* \in \DD$ is the
Normal distribution $\NN(\mu_f(\bx_*), \sigma_f(\bx_*)^2 + \sigmaobs^2)$
where $\mu_f(\cdot)$ and $\sigma_f(\cdot)^2$ are given by
\begin{align}
\label{eqn:gppredmean}
    \mu_f(\bx_*) &= {\bK_{*,N}}^{\rm T}  {(\bK_{N,N} + \sigmaobs^2 \mathbb{1}_N)}^{-1}\by  \\
\label{eqn:gppredvar}
    \sigma_f(\bx_*)^2 &=  K_{**} - {\bK_{*,N}}^{\rm T}  {(\bK_{N,N} + \sigmaobs^2 \mathbb{1}_N)}^{-1}    {\bK_{*,N}}
\end{align}
Here $K_{**} =  K(\bx_*, \bx_*) $ and ${\bK_{*,N}}$ is the column vector with elements $({\bK_{*,N}})_n=K(\bx_*, \bx_n)$. 

\section{Nearest neighbor cross-validation losses}
\label{sec:cv}

The marginal log likelihood of a GP can be written as a sum of logs of univariate posterior conditionals, where
for a fixed, arbitrary ordering $\{1, ..., N\}$ we have
\begin{equation}
\label{eqn:lmll}
\LLMLL \equiv \log p(\by | \bX ) = \Sigma_{n=1}^N \log p(y_n | \by_{<n}, \bx_{\le n})
\end{equation}
 and where $\by_{<n} \equiv \{y_1, ... , y_{n-1} \}$ and  $\bx_{\le n} \equiv \{\bx_1, ... , \bx_{n} \}$.\footnote{We
 define $\by_{<1} \equiv \emptyset$. Note that here and elsewhere we suppress dependence on the kernel hyperparameters.}
 Summing over all $N!$ permutations $\tau \in S^N$ we obtain
\begin{equation}
\LLMLL = \frac{1}{N!}\sum_{\tau \in S^N}\sum_{n=1}^N \log p(y_n^\tau | \by_{<n}^\tau, \bx_{\le n}^\tau) \equiv \sum_{n=0}^{N-1} \LLCV^n
\label{eqn:sigmamll}
\end{equation}
where $\tau$ superscripts denote application of the permutation $\tau$.\footnote{For example
$y_n^\tau \equiv y_{\tau(n)}$ and $\by_{<n}^\tau \equiv \{y_{\tau(1)}, ... , y_{\tau(n-1)} \} $}
Additionally on the RHS of Eqn.~\ref{eqn:sigmamll} we have grouped the $N \! \times \! N!$
terms in the sum w.r.t.~the number of data points that each term conditions on,
i.e.~each term in $\LLCV^n$ conditions on exactly $n$ data points and is implicitly defined by Eqn.~\ref{eqn:sigmamll}.
 For example we have
\begin{equation}
    \begin{split}
    \LLCV^{N-1} &=  \frac{1}{N} \sum_{n=1}^N \log p(y_n | \by_{-n}, \bX) \\
    \LLCV^{0} &=   \frac{1}{N}\sum_{n=1}^N \log p(y_n | \bx_n)
\label{eqn:cvN}
    \end{split}
\end{equation}
 where $\by_{-n} \equiv \{ y_1, ..., y_{n-1}, y_{n+1}, ..., y_N \}$.
  This decomposition makes a number of properties of the marginal log likelihood apparent.
First, the MLL is directly linked to the average posterior predictive performance conditioned on \emph{all} possible
training data sets, including the empty set \citep{fong2020marginal}. Second, the inclusion of conditioning sets
that are only a small fraction of the full dataset---in the extreme case empty conditioning sets as in $\LLCV^{0}$---means
that $\LLMLL$ can exhibit substantial dependence on the prior. Indeed the term $\LLCV^{0}$ scores the model exclusively
with respect to~the \emph{prior} predictive. Conversely, $\LLCV^{N-1}$ exhibits the least dependence on the prior.

In the following we will use $\LLCV^{N-1}$ as the basis for our training objective, in
particular using it to learn the hyperparameters of the kernel.
This choice is motivated by two observations.
First, as we have just argued, we expect $\LLCV^{N-1}$ to provide robustness against prior mis-specification due to its
reduced dependence on the prior. Indeed it is well known that conventional Bayesian inference can be suboptimal when
the model is mis-specified: see \citep{masegosa2019learning} and references therein for recent discussion.
Second, while $\LLCV^{N-1}$ depends on the entire dataset, \emph{individual} predictive distributions
$p(y_n | \by_{-n}, \bX)$ in the sum typically exhibit \emph{non-negligible} dependence on only a
small subset of the conditioning data.\footnote{At least for approximately compactly-supported kernels
like RBF or Mat{\'e}rn. Our method is not immediately applicable to, e.g., periodic kernels as are commonly used in
time-series applications.} This latter observation opens the door to nearest neighbor truncation,
which we describe next.

\subsection{Nearest neighbor truncation}
\label{sec:nn}

In the regression case (see Eqn.~\ref{eqn:gppredmean}-\ref{eqn:gppredvar})
 computing  $\LLCV^{N-1}$ in Eqn.~\ref{eqn:cvN} has $\OO(N^4)$ cost if done naively,
which is very expensive for $N \gtrsim 10^3$.
While this can be reduced to $\OO(N^3)$ for a stochastic estimate or
if care is taken with the algebra \citep{petit2020towards, ginsbourger2021fast}, this still precludes a training algorithm
that scales to large datasets with millions of data points. To enable scalability, we apply a $k$-nearest-neighbor truncation to  $\LLCV^{N-1}$
to obtain the $k$-truncated leave-one-out (\textbf{LOO-}$\mathbf{k}$) objective 
\begin{tcolorbox}[enhanced,ams align,height=44pt,width=0.95\linewidth,center,colback=gray!7!white]
 \LLL^k \equiv \frac{1}{N} \sum_{n=1}^N \log p(y_n | \by_{n, k}, \bX_{n,k}, \bx_n)
\label{eqn:cvk}
\end{tcolorbox}
 where we use $k=\infty$ to denote the non-truncated objective.
 Here the pair $\left( \by_{n, k}, \bX_{n, k} \right)$ denotes the $k$-nearest-neighbors of $\bx_n$ (and the corresponding targets) as determined
using the Euclidean metric defined with kernel length scales ${\rho_i}$.\footnote{That is we compute nearest neighbors w.r.t.~the distance function
 $d(\bx, \bz) = \sqrt{\sum_i (x_i - z_i)^2 / \rho_i^2}$. Note that by definition $\bX_{n, k}$ does \emph{not} contain $\bx_n$.}
For univariate regression computing Eqn.~\ref{eqn:cvk} has a $\OO(Nk^3)$ cost.
Utilizing data subsampling to obtain a stochastic estimate $\LLLst$, this cost becomes $\OO(Bk^3)$ for mini-batch size $B$.
Notably, the bottleneck in computing $\LLLst$ involves a batch
Cholesky decomposition of a $B \times k \times k$ tensor, which
can be done extremely efficiently on a GPU for $k \lesssim 500$ even for $B \sim 100$.
The nearest neighbor search takes $\mathcal O( N \log N)$ time using standard algorithms like k-d trees \citep{bentley1975multidimensional},
though for $N \lessapprox 10^7$ it is often faster to use $\mathcal O( N^2 )$ brute force algorithms that make use of GPU parallelism.\footnote{
  Modern nearest neighbor libraries, e.g.~FAISS \citep{johnson2019billion}, perform nearest neighbor searches in a map-reduce fashion.
  This yields a $\mathcal O(N)$ memory requirement, which is feasible on a single GPU for $N \lessapprox 10^7$.
  For larger datasets it is possible to achieve speed-ups with quantization,
  trading off speed for accuracy.
}
In practice we recompute the nearest neighbor index somewhat infrequently, e.g.~after every $50^{\rm th}$ gradient update.
See Algorithm~\ref{algo} in the supplement for a summary of the training procedure. Note that since there
are no inducing points, the sole purpose of the training step is to identify good kernel hyperparameters.

To make a prediction for a test point $\bx_*$ we simply compute the $k$-truncated posterior predictive
distribution $p(y_n | \by_{*, k}, \bX_{*,k}, \bx_*)$. Here the nearest neighbors $(\by_{*, k}, \bX_{*,k})$
are formed using a pre-computed nearest neighbor index that only needs to be computed once.
The nearest neighbor query takes $\OO(N)$ time using a brute force index and $\OO(\log N)$ time for a k-d tree.
The cost of forming the predictive distribution is then $\OO(k^3)$.
We note that the non-parametric nature of our predictive distribution means that the training data must
be available at test time.

\subsection{Binary classification}
\label{sec:class}

Above we implicitly assume that the LOO posterior probability $p(y_n | \by_{n, k}, \bX_{n,k}, \bx_n)$ can be computed
analytically. What if this is not the case, as happens in classification?
In this section we briefly describe how we define a LOO training objective for binary classification.
The basic strategy is to introduce auxiliary variables that restore Gaussianity and thus enable closed
form (conditional) posterior distributions.
Let $\DD = \{(\bx_n, y_n)\}_{n=1}^N$ with $y_n \in \{-1, 1\}$
%
 and consider a GP classifier with likelihood $p(y_n | f(\bx_n)) = \left[1 + \exp(-y_n f(\bx_n))\right]^{-1}$ governed by a logistic function.
 We introduce a $N$-dimensional vector of P\olya-Gamma  \citep{polson2013bayesian} auxiliary variables $\bomega$ and exploit the identity
 $\left[1 + e^\psi\right]^{-1} =   \tfrac{1}{2}  \EE_{p(\omega | 1, 0)} \left[  \exp \left(-\tfrac{1}{2} \omega \psi^2 -\tfrac{1}{2}\psi \right) \right] $
  to massage the likelihood terms
 into Gaussian form and end up with a joint density $p(\by, \bff, \bomega | \bX)$ proportional to 
\begin{equation}
    p(\bomega) p(\bff | \bK_{N,N}) \exp\left(\tfrac{1}{2} \by^{T} \bff - \tfrac{1}{2}  \bff^{T} \bOmega \bff  \right)
\end{equation}
 where $\bOmega \equiv {\rm diag}(\bomega)$ is a diagonal $N \times N$ matrix. Note that this augmentation is exact.
 We proceed to integrate out $\bff$ and perform variational inference w.r.t.~$\bomega$. This results in the variational objective
\begin{equation}
\LL_{\rm ELBO} = \EE_{q(\bomega)} \left[ \log p(\by | \bX, \bomega) \right] - {\rm KL}(q(\bomega) | p(\bomega) )
\end{equation}
 where we take $q(\bomega)$ to be a mean-field log-Normal variational distribution and KL denotes the Kullback-Leibler divergence.
 We then replace $\log p(\by | \bX, \bomega)$  with its LOO approximation to obtain
\begin{equation}
  \begin{split}
    \LLL &\equiv \tfrac{1}{N} \Sigma_{n=1}^N \EE_{q(\bomega)} \left[  \log p(y_n | \by_{-n}, \bX, \bomega_{-n}) \right]
    \\
    &\phantom{=} \:\: - {\rm KL}(q(\bomega) | p(\bomega) ) 
    \label{eqn:loobin}
  \end{split}
\end{equation}
where the (conditional) posterior predictive distribution $p(y_n | \by_{-n}, \bX, \bomega_{-n})$ is given by
 \begin{equation}
     p(y_n | \by_{-n}, \bX, \bomega_{-n}) \equiv {\textstyle \int} df_n \; p(y_n | f_n) p(f_n | \by_{-n}, \bX, \bomega_{-n})
\label{eqn:postpredyn}
\end{equation}
Here the (conditional) posterior over the latent function value $f_n$, namely $p(f_n | \by_{-n}, \bX, \bomega_{-n})$, is given by the Normal distribution
$\NN(f_n  | \mu_n, \sigma_n^2)$
with mean and variance equal to
\begin{align*}
\mu_n &= 
             \tfrac{1}{2} \bK_{-n, n}^{\rm T} \left( \bK_{-n, -n}+ \bOmega_{-n}^{-1} \right)^{-1} \bOmega_{-n}^{-1} \by_{-n} \\
\sigma_n^2 &=  K_{n,n}-\bK_{-n, n}^{\rm T} \left( \bK_{-n,-n}+ \bOmega_{-n}^{-1} \right)^{-1} \bK_{-n, n} \\
             \bK_{-n, n} &\equiv K(\bX_{-n}, x_n)  \in \RR^{n-1,1} \\ 
    \bK_{-n,-n} &\equiv K(\bX_{-n}, \bX_{-n}) \in \RR^{n-1,n-1}
\end{align*}
 We approximate the univariate integral in Eqn.~\ref{eqn:postpredyn} with Gauss-Hermite quadrature.
 Our final objective function is obtained by applying a $k$-nearest-neighbor truncation to Eqn.~\ref{eqn:loobin}.
 To maximize this objective function we use standard techniques from stochastic variational inference, including data subsampling
 and reparameterized gradients of $\bomega$.
 At test time predictions can be obtained by computing Eqn.~\ref{eqn:postpredyn} after
 conditioning on a sample $\bomega \sim q(\bomega)$. Note that in practice we use a single sample, 
 since we found negligible gains from averaging over multiple samples.
 The computational cost of a training iteration is $\OO(Bk^3 + BQ)$, where $Q$ is the number of points in the
 quadrature rule. 
 For a more comprehensive derivation and additional details on binary classification see Sec.~\ref{app:binclass}.

\begin{figure}[t!]
  \centering
  \includegraphics[width=0.77\linewidth]{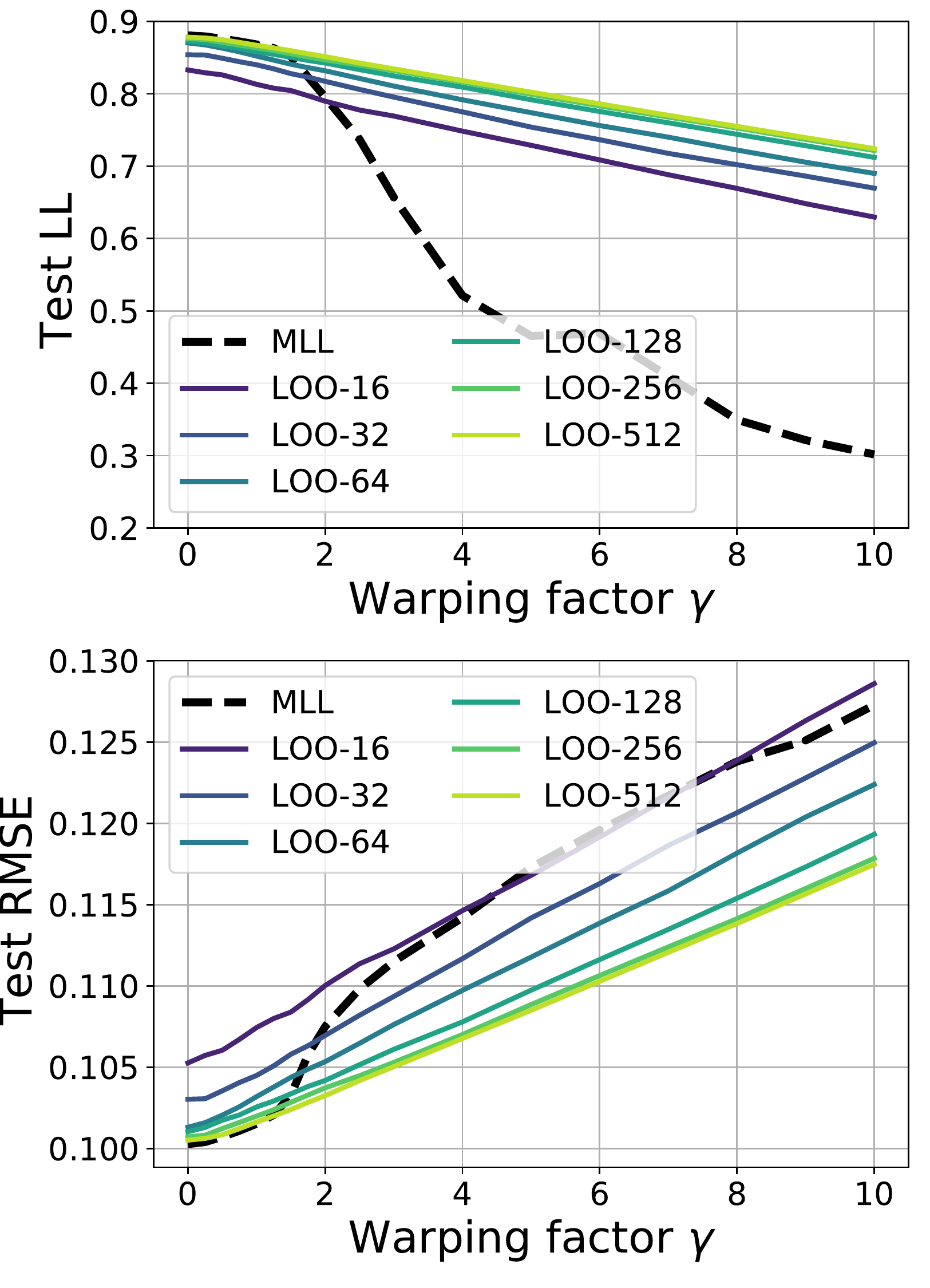}
    \caption{We compare predictive performance of a GP trained with a LOO-$k$ objective (see Eqn.~\ref{eqn:cvk}) to a GP
    trained via MLL (marginal log likelihood), where the GP regressor is mis-specified due
  to non-stationarity in the dataset introduced by a warping function controlled by $\gamma$.
    As $\gamma$ increases and the dataset becomes more non-stationary,
    the LOO-$k$ GP exhibits superior predictive performance, both w.r.t.~log likelihood (LL) and root mean squared error (RMSE).
    See Sec.~\ref{sec:misexp} for details.}
    \label{fig:mis}
\end{figure}

\subsection{Multi-class classification}
\label{sec:multiclass}

The P\olya-Gamma auxiliary variable construction used in Sec.~\ref{sec:class} can be generalized to the multi-class setting with $K$ classes using
a stick-breaking construction \citep{linderman2015dependent}.
Among other disadvantages, this construction requires choosing a class ordering.
To avoid this, and to obtain linear computational scaling in the number of classes $K$,
we instead opt for a one-against-all construction,
which results in a $\OO(BKk^3 + BKQ)$  computational cost  per training iteration.
We refer the reader to Sec.~\ref{app:multiclass} for details and Sec.~\ref{sec:multiexp} for empirical results.

\subsection{Other likelihoods}
\label{sec:other}

The auxiliary variable construction in Sec.~\ref{sec:class}-\ref{sec:multiclass} can be extended to a number of
other likelihoods, including binomial and negative binomial likelihoods as well as Student's t likelihood.
See Sec.~\ref{app:other} for more details.

\section{Related work}
\label{sec:related}

\begin{figure}[t!]
  \centering
  \includegraphics[width=1.05\linewidth]{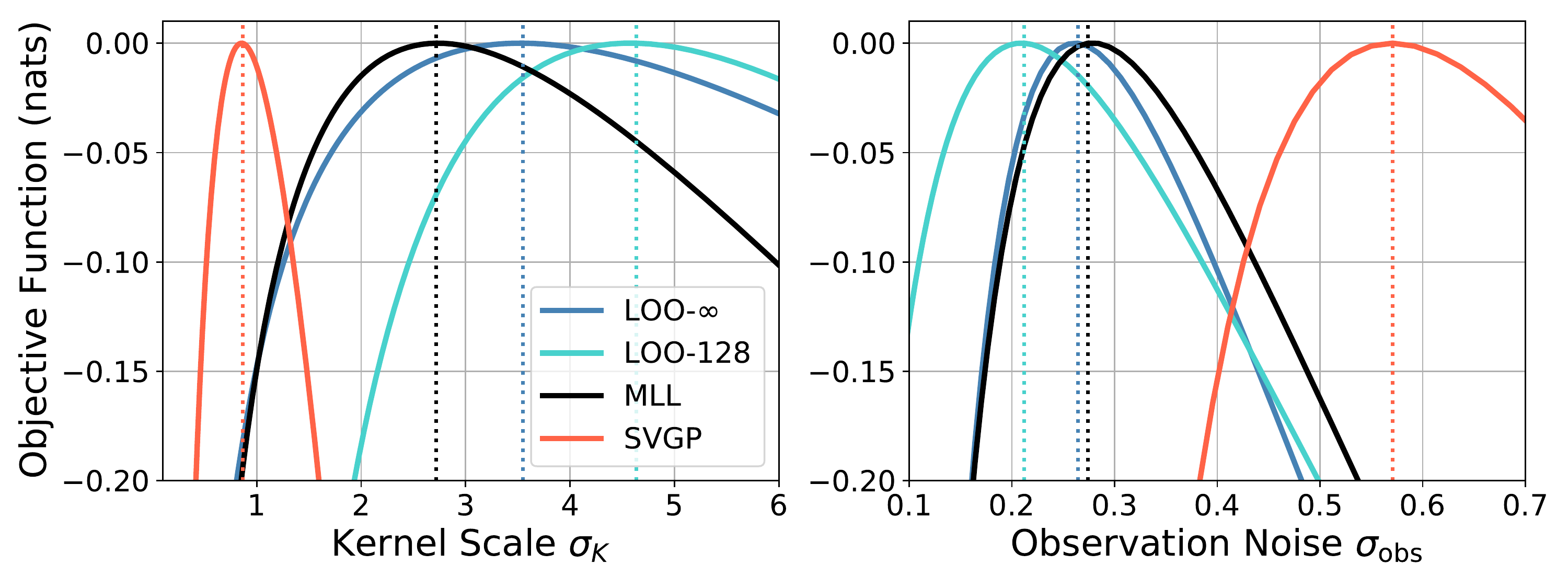}
  \caption{We depict how the objective functions for MLL, LOO-$k$ and SVGP vary
           as a function of the kernel scale $\sigma_K$ and the observation noise $\sigmaobs$ for
           the UCI Bike dataset. To ease comparison we shift each objective function so that its maximum value is zero.
           $\sigma_K$ and $\sigmaobs$ are maximized at $\{{\scriptstyle 3.55, 4.64, 2.72, 0.86 }\}$
           and $\{{\scriptstyle 0.26, 0.21, 0.27, 0.57}\}$ for LOO-$\infty$, LOO-$128$, MLL, and SVGP, respectively.
           Each objective is normalized by the training dataset size $N$.
           See Sec.~\ref{sec:objexp} for details.
           }\label{fig:hypercurve}
%
%
%
\end{figure}

Our work is related to various research directions in machine learning and statistics.
Here we limit ourselves to an abbreviated account and refer the reader to Sec.~\ref{app:related} for a more detailed discussion.

Nearest neighbor constructions in the GP context have been explored by several authors.
\citet{datta2016hierarchical} define a Nearest Neighbor Gaussian Process, which is a valid stochastic process,
derive a custom Gibbs inference scheme, and illustrate their approach on geospatial data.
Vecchia approximations \citep{vecchia1988estimation,katzfuss2021general} exploit a nearest neighbor approximation
to the MLL. Both these approaches can work well in 2 or 3 dimensions but tend to struggle in higher dimensions due to the
need to choose a fixed ordering of the data. 
\citet{tran2021sparse} introduce a variational scheme for GP inference (SWSGP) that leverages nearest neighbor
truncation within an inducing point construction.
\citet{gramacy2015local} introduce
a `Local Gaussian Process Approximation' for regression that iteratively constructs a nearest neighbor conditioning set
at test time; in contrast to our approach there is no training phase.

\begin{figure}[t!]
  \centering
  \includegraphics[width=1.05\linewidth]{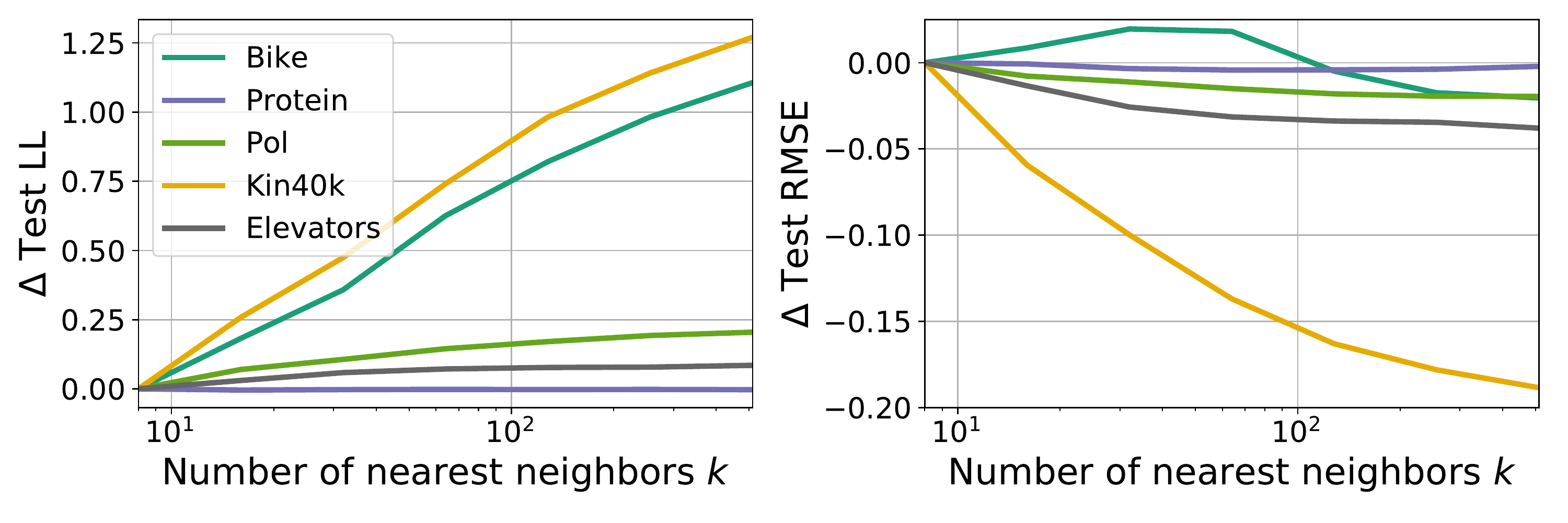}
  \caption{We depict how the predictive performance of a LOO-$k$ GP varies w.r.t.~the number of
    nearest neighbors $k$. For each of the five UCI datasets, we depict mean test log likelihood
    and mean RMSE averaged over 10 train/test splits. To ease comparison we shift each LL/RMSE curve
    so that it intersects zero at $k=8$.
           See Sec.~\ref{sec:k} for discussion.
           }\label{fig:k}
\end{figure}

Cross-validation (CV) in the GP (or rather kriging) context was explored as early as 1983 by \citet{dubrule1983cross}.
\citet{bachoc2013cross} compares predictive performace of GPs fit with MLL and CV and concludes
that CV is more robust to model mis-specification. 
\citet{smith2016differentially} consider CV losses in the context of differentially private GPs.
Recent work explores how the CV score can be efficiently computed for GP regressors
\citep{ginsbourger2021fast,petit2020towards}.
\citet{jankowiak2020parametric} introduce an inducing point approach for GP regression, PPGPR, that like our approach 
uses a loss function that is defined in terms of the predictive distribution. Indeed our approach
can be seen as a non-parametric analog of PPGPR, and Eqn.~\ref{eqn:sigmamll} provides a novel
conceptual framing for that approach.
\citet{fong2020marginal} explore the connection between MLL and CV in the context
of model evaluation and consider a decomposition like that in Eqn.~\ref{eqn:sigmamll} 
specialized to the case of exchangeable data. They also advocate
using a Bayesian \emph{cumulative} leave-P-out CV score for fitting models, although the computational
cost limits this approach to small datasets.

Various approaches to approximate GP inference are reviewed in \citet{liu2020gaussian}.
An early application of inducing points is described in \citet{snelson2006sparse},
which motivated various extensions to variational inference \citep{titsias2009variational,hensman2013gaussian,hensman2015scalable}.
\citet{wenzel2019efficient,galy2020multi} introduce an approximate inference scheme
for binary and multi-class classification that exploits inducing points, P\olya-Gamma auxiliary variables, and variational inference.


\section{Experiments}
\label{sec:exp}

In this section we present an empirical evaluation of GPs trained with the LOO-$k$ objective
Eqn.~\ref{eqn:cvk}. In Sec.~\ref{sec:misexp}-\ref{sec:degenerate} we explore general characteristics of our method
and in Sec.~\ref{sec:uniexp}-\ref{sec:multiexp} we compare our method to a variety of baseline methods for
GP regression and classification.

\begin{figure}[t!]
  \centering
  \includegraphics[width=\linewidth]{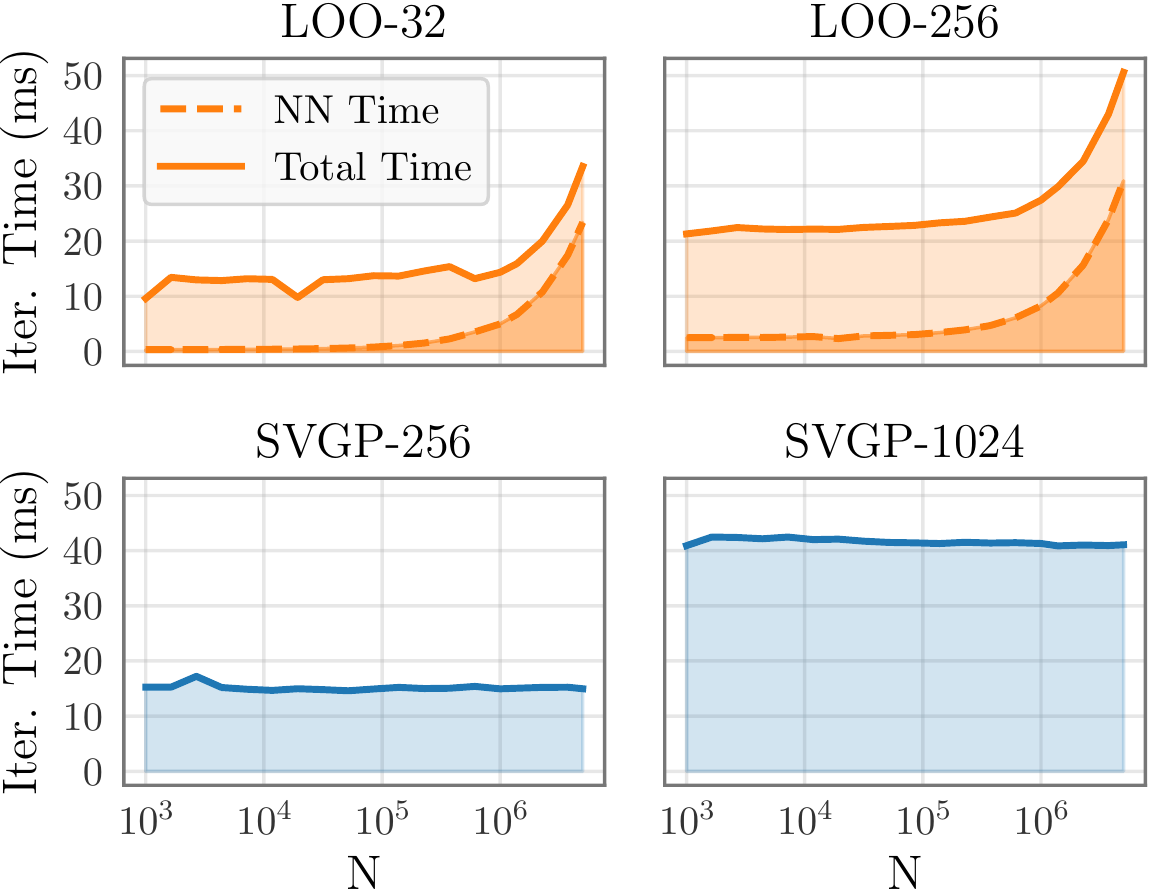}
  \caption{We compare the time per training iteration for LOO-$k$  ($k=32$ and $k=256$ nearest neighbors)
    versus SVGP ($M=256$ and $M=1024$ inducing points) on a univariate regression task in $D=25$ dimensions.
    Along the horizontal axis we vary the size of the training dataset $N$.
    We use a mini-batch size of 128 and run on a GeForce RTX 2080 GPU.
    The dotted line indicates the time spent on nearest neighbor queries; we update the nearest neighbor
    index every 50 gradient steps.
    }\label{fig:runtime}
\end{figure}

\subsection{Model mis-specification}
\label{sec:misexp}

We explore whether the LOO-$k$ objective in Eqn.~\ref{eqn:cvk} is robust to model mis-specification,
as we would expect following the discussion in Sec.~\ref{sec:cv}.
Since models can be mis-specified in a great variety of ways,
it is difficult to make quantitative statements about mis-specification in general terms.
Instead we choose a simple controlled setting where we can toggle the degree of mis-specification.
First we sample $8096$ data points $\bX$ uniformly from the cube $[-1, 1]^4 \subset \RR^4$.
Next we generate targets $\by$ using a GP prior specified by an isotropic stationary RBF kernel with observation noise $\sigmaobs=0.1$.
We then apply a coordinatewise warping to each $\bx_n$, where the warping is given by the identity
mapping for $x_i \ge 0$ and $x_i \rightarrow (1 + \gamma)x_i$ for $x_i <0$.
We use half the data points for training and the remainder for testing predictive performance.
The results can be seen in Fig.~\ref{fig:mis}.
We see that as $\gamma$ increases and the dataset becomes more non-stationary,
the LOO-$k$ GP exhibits superior predictive performance.
We note that the degraded test log likelihood of the MLL GP
for large $\gamma$ stems in part from severely underestimating the observation noise $\sigmaobs$.
Fig.~\ref{fig:mis} confirms our expectation that using a CV-based objective
can be more robust to model mis-specification, especially as the latter becomes more severe.

\begin{figure*}[t!]
  \centering
  \includegraphics[width=\linewidth]{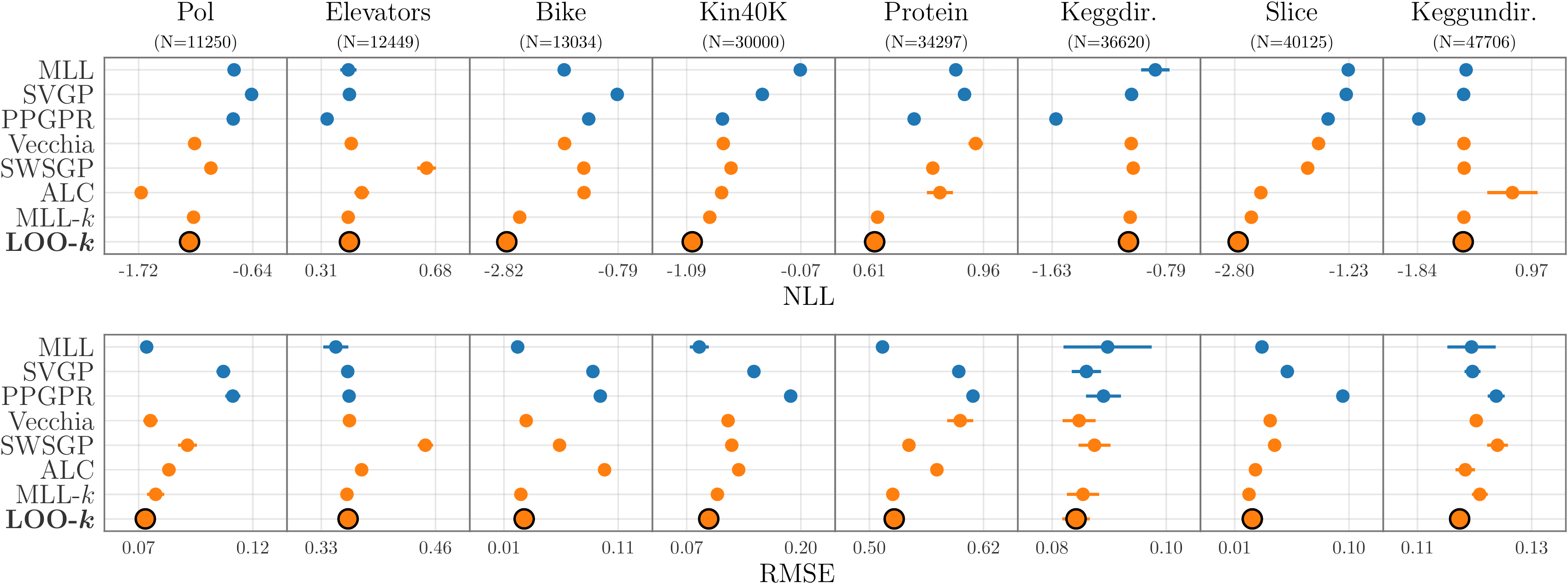}
  \caption{We depict predictive negative log likelihoods (NLL, top) and root mean squared error (RMSE, bottom) for 8
    univariate regression datasets, comparing LOO-$k$ to 7 baseline methods.
    Results are averaged over 10 dataset splits.
    Lower numbers are better. Methods in orange utilize nearest neighbors.
    Here and elsewhere horizontal bars denote standard errors.
    Results for SVGP, PPGPR, and MLL are reproduced from \citep{jankowiak2020parametric}.}
  \label{fig:uci_regression}
\end{figure*}

\subsection{Objective function comparison}
\label{sec:objexp}

In Fig.~\ref{fig:hypercurve} we compare how the MLL, SVGP, and LOO-$k$ objective functions depend
on the kernel hyperparameters $\sigma_K$ and $\sigmaobs$.
First we subsample the UCI Bike dataset to $N=2000$ datapoints. We then train a GP using MLL on this dataset
and keep all the hyperparameters fixed to their MLL values except for the one hyperparameter that is varied.
Fig.~\ref{fig:hypercurve} makes apparent the well-known tendency of SVGP to overestimate the observation
noise $\sigmaobs$ and, consequently, prefer a smaller value of $\sigma_K$ \citep{bauer2016understanding}.
This overestimation of $\sigmaobs$ can lead to severe overestimation of uncertainty at test time.
Since $\sigma_K$ and $\sigmaobs$ appear symmetrically in Eqn.~\ref{eqn:gppredvar}
and likewise in $\LLL^k$, the LOO-$k$ GP does not exhibit the same tendency
(as argued in \citep{jankowiak2020parametric}).

\subsection{Dependence on number of nearest neighbors $k$}
\label{sec:k}

In Fig.~\ref{fig:k} we explore how predictive performance of a LOO-$k$ GP depends on the number of
nearest neighbors $k$. As expected, we generally find that performance improves as $k$ increases, although
the degree of improvement depends on the particular dataset.
In addition the marginal gains of increasing $k$ past $k \sim 128$ are small on most datasets.
This finding is advantageous for our method with respect to computational cost, whereas inducing point methods often
require $M \gtrsim 500$ to achieve good model fit.
This tendency is easy to understand, since the $M$ inducing points need to `compress' the entire
dataset, while the $k$ nearest neighbors only need to model the vicinity of a given test point.

\subsection{Runtime performance}
\label{sec:runtime}

In Fig.~\ref{fig:runtime} we compare the runtime performance of LOO-$k$ to SVGP on datasets from $N=1000$ to $N=5\text{ million}$.
As discussed further in Sec.~\ref{app:complexity}, we can take these two methods' runtimes as
representive of other nearest neighbor (e.g.~Vecchia) and inducing point (e.g.~PPGPR) methods, respectively.
Thanks to highly parallel GPU-accelerated nearest neighbor algorithms \citep{johnson2019billion}, only a small fraction of LOO-$k$
training time is devoted to nearest neighbors queries up to $N \sim 10^6$.
Though these queries become more costly as $N$ approaches $10$ million,
LOO-$k$ is comparable to SVGP in this regime, and search time can be improved by sharding or other approximations.
Since a LOO-$k$ regressor is fully parameterized by a handful
of kernel hyperparameters and does not make use of variational parameters, it requires substantially fewer gradient steps to converge.
For example on the Kegg-undirected dataset with $N=47706$ considered in the next section we find that LOO-256 trains $\sim\! 4$x faster than SVGP with $M=1024$ inducing points.

\subsection{Degenerate data regime}
\label{sec:degenerate}

To better understand the limitations of LOO-$k$ we run an experiment in which we create artificially `degenerate'
datasets by adding a noisy replicate of each data point in the training set
(adding $\mathcal N(0, 10^{-4})$ noise both to inputs $\mathbf x$ and responses $y$).
We expect that a potential failure mode of nearest neighbor methods like LOO-$k$ is a reduced ability to model
long-distance correlations. Indeed we expect better performance from a global inducing point method like SVGP in
this regime.
As expected---see Table~\ref{table:degenerate} in the supplement for complete results---the performance of SVGP-512 is not much affected by the addition of severe degeneracy,
while LOO-64 exhibits a large loss in performance on 3/4 datasets
(although LOO still exhibits better predictive performance than SVGP on 2/4 degenerate datasets).
Ultimately this loss in performance in LOO-$64$ can be traced to a systematic preference for smaller kernel lengthscales.
While the good empirical results on most datasets in subsequent sections suggest that many datasets do not
exhibit such degeneracy, this limitation of LOO-$k$ should be kept in mind.

\begin{figure}[t!]
  \centering
  \includegraphics[width=0.99\linewidth]{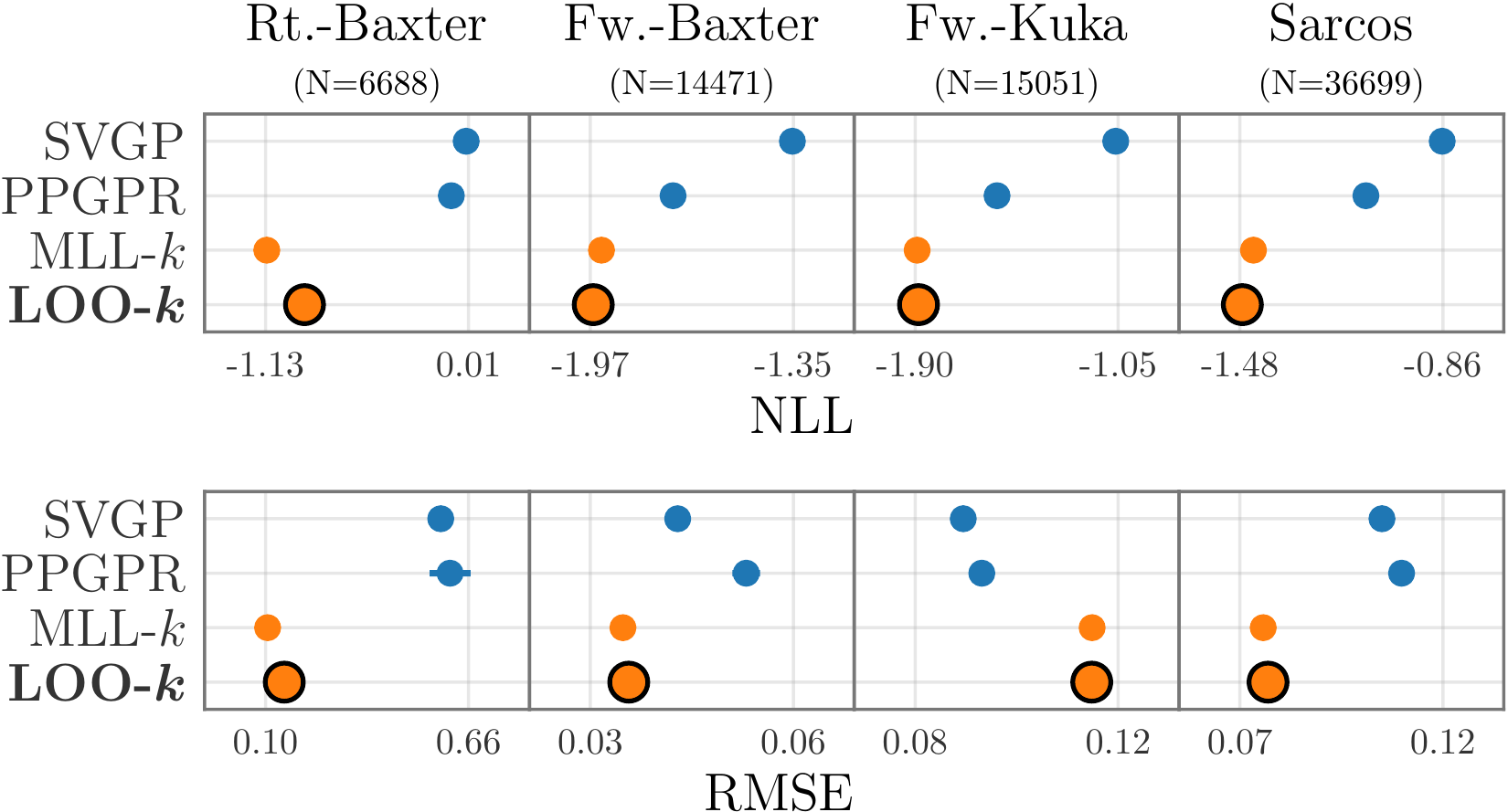}
  \caption{We depict predictive negative log likelihoods (NLL, top) and root mean squared error (RMSE, bottom) for 4 multivariate regression datasets.
    Results are averaged over 10 dataset splits.
    Lower numbers are better.
    Results for SVGP and PPGPR are from \citep{jankowiak2020deep}.}
  \label{fig:multi_regression}
\end{figure}

\subsection{Univariate regression}
\label{sec:uniexp}

We compare the performance of GP regressors trained with the LOO-$k$ objective
in Eqn.~\ref{eqn:cvk} to six scalable baseline methods.
SVGP \citep{hensman2013gaussian} and PPGPR \citep{jankowiak2020parametric}
are both inducing point methods; SVGP targets the MLL via an ELBO lower bound, while PPGPR uses a regularized cross-validation loss.
The remaining baselines, MLL-$k$, Vecchia, SWSGP, and ALC, exploit nearest neighbors.
MLL-$k$ uses a biased $k$-nearest-neighbor truncation of the MLL for training \citep{chen2020stochastic}.
Vecchia \citep{vecchia1988estimation} uses a different $k$-nearest-neighbor approximation of the MLL that requires specifying a fixed ordering of the data.
SWSGP \citep{tran2021sparse} combines nearest neighbor approximations and inducing point methods,
where each data point only depends on a subset of inducing points.
ALC \citep{gramacy2015local} iteratively constructs a nearest neighbor conditioning set
at test time using a variance reduction criterion and chooses hyperparameters using the `local' MLL.
See Sec.~\ref{sec:conceptual} in the supplemental materials for a conceptual framing of these different approaches.
We also include a comparison to GPs trained with MLL using the methodology in \citet{wang2019exact}.
See Sec.~\ref{app:exp} for additional experimental details.

The results are depicted in Fig.~\ref{fig:uci_regression} and summarized in Table~\ref{tab:uci_regression_rank}.
We find that LOO-$k$ exhibits the best predictive performance overall, both w.r.t.~log likelihood and RMSE,
followed by MLL-$k$.
ALC performs well on some datasets but poorly on others; indeed we are unable to obtain reasonable
results on the Kegg-directed dataset. Vecchia does poorly overall, presumably due to the need to use a strict ordering of
the data. SVGP also does poorly overall; as argued by \citet{bauer2016understanding} and \citet{jankowiak2020parametric}
degraded performance w.r.t.~log likelihood can be traced to a tendency to overestimate the observation noise and consequently
underestimate function uncertainty. PPGPR exhibits good log likelihoods but poor RMSE performance due to the priority
it places on uncertainty quantification.
It is also worth highlighting that the nearest neighbor methods can perform well in high-dimensional input spaces;
e.g.~the Slice dataset is 380-dimensional.

\begin{table}
  \centering
  \caption{
      \hspace{-1.10em}
    We summarize the performance ranking of various GP methods on the univariate regression experiments in Fig.~\ref{fig:uci_regression}
    averaged over all dataset splits.
    CRPS is the Continuous Ranking Probability Score \citep{gneiting2007strictly}.
    Lower is better for all metrics.
  }
  \label{tab:uci_regression_rank}
  \resizebox{0.99\linewidth}{!}{%
    \begin{tabular}{ccccccccc}
\toprule
{} &   &    SVGP &   PPGPR & Vecchia &   SWSGP &     ALC & MLL-$k$ &            LOO-$k$ \\
Value &           &         &         &         &         &         &         &                    \\
\midrule
NLL   &   &  $5.60$ &  $3.51$ &  $4.64$ &  $5.16$ &  $4.41$ &  $2.64$ &  $\mathbf{ 2.04 }$ \\
RMSE  &   &  $4.59$ &  $6.21$ &  $3.44$ &  $4.76$ &  $4.67$ &  $2.34$ &  $\mathbf{ 1.99 }$ \\
CRPS  &   &  $5.78$ &  $4.09$ &  $4.69$ &  $4.74$ &  $4.06$ &  $2.65$ &  $\mathbf{ 2.00 }$ \\
\bottomrule
\end{tabular}

  }
\end{table}

\subsection{Multivariate regression}
\label{sec:multivarexp}

\begin{figure*}[t!]
  \centering
  \includegraphics[width=1.03\linewidth]{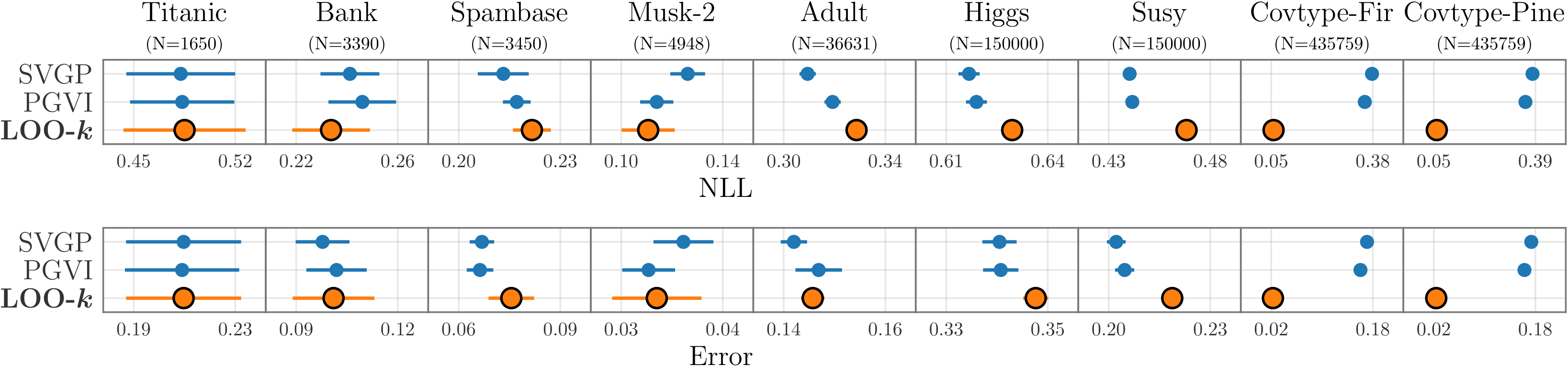}
  \caption{
    We depict predictive negative log likelihood (NLL, top) and error (bottom) for 9 binary classification datasets.
    Results are averaged over 5 dataset splits.
    Lower numbers are better.
  }
  \label{fig:binary_classification}
\end{figure*}

We continue our empirical evaluation by considering four multivariate regression datasets.
In each dataset the input and output dimensions correspond to various joint positions/velocities/etc.~of a robot.
Each GP regressor employs the structure of the linear model of coregionalization (LMC) \citep{alvarez2012kernels}.
We compare against three baseline methods: SVGP, PPGPR, and MLL-$k$.
The results are depicted in Fig.~\ref{fig:multi_regression} and summarized in Table~\ref{tab:multi_regression_rank}.
We find that the two nearest neighbor methods, LOO-$k$ and MLL-$k$, outperform the two inducing point methods,
SVGP and PPGPR, with LOO-$k$ and MLL-$k$ performing the best w.r.t.~NLL and RMSE, respectively.
Strikingly, for 3/4 datasets
the RMSEs for the nearest neighbor methods are much smaller than for the inducing point methods, even
though we use no more than $k=32$ nearest neighbors.
We hypothesize that it is difficult to capture the complex non-linear dynamics underlying these datasets
using a limited number of inducing points. In addition, training objectives that depend on large numbers of inducing points
can be challenging to optimize, potentially resulting in suboptimal solutions.
This highlights one of the advantages of methods like LOO-$k$ and MLL-$k$,
which avoid  optimization in input space and
 result in flexible non-parametric nearest neighbor predictive distributions.

\begin{table}
{\caption{
    We summarize the performance ranking of various GP methods on the multivariate regression experiments
    in Fig.~\ref{fig:multi_regression} averaged over all dataset splits.
}\label{tab:multi_regression_rank}}
{ 
  \resizebox{0.85\linewidth}{!}{%
    \begin{tabular}{cccccc}
\toprule
{} &   &    SVGP &   PPGPR &            MLL-$k$ &            LOO-$k$ \\
Value &           &         &         &                    &                    \\
\midrule
NLL   &   &  $3.94$ &  $3.06$ &             $1.57$ &  $\mathbf{ 1.43 }$ \\
RMSE  &   &  $2.61$ &  $3.39$ &  $\mathbf{ 1.80 }$ &             $2.20$ \\
\bottomrule
\end{tabular}

  }
} 
\end{table}

\begin{table}[h!]
{\caption{We summarize the performance ranking of various GP methods for binary classification
    averaged over all dataset splits. Lower is better for both metrics.}\label{tab:binary_classification_rank}}
{ 
  \vspace{-1.0mm}
  \resizebox{0.65\linewidth}{!}{%
    \begin{tabular}{ccccc}
\toprule
{} &   &               SVGP &               PGVI & LOO-$k$ \\
Value &           &                    &                    &         \\
\midrule
NLL   &   &  $\mathbf{ 1.89 }$ &             $2.04$ &  $2.07$ \\
Error &   &             $1.99$ &  $\mathbf{ 1.88 }$ &  $2.13$ \\
\bottomrule
\end{tabular}

  }
} 
\end{table}

\subsection{Binary classification}
\label{sec:binexp}

Next we compare the predictive performance of the LOO-$k$ objective for binary classification, Eqn.~\ref{eqn:loobin},
to two scalable GP baselines. Both SVGP \citep{hensman2015scalable} and PGVI \citep{wenzel2019efficient} are inducing point
methods that utilize variational inference; PGVI also makes use of natural gradients and P\olya-Gamma augmentation.
The results are depicted in Fig.~\ref{fig:binary_classification}
and summarized in Table~\ref{tab:binary_classification_rank}. The predictive performance is broadly comparable for most datasets.
Most striking is the superior performance of LOO-$k$ on the two Covtype datasets, which exhibit (geospatial) decision boundaries
with complex topologies that are difficult to capture with a limited number of inducing points.

\subsection{Multi-class classification}
\label{sec:multiexp}

We compare the predictive performance of the LOO-$k$ objective for multi-class classification to a SVGP baseline.
We consider 13 datasets with 5 splits per dataset for
a total of 65 splits. We find that LOO-$k$ outperforms SVGP on 41/65 and 42/65 splits
w.r.t.~log likelihood and error, respectively. See Sec.~\ref{app:addfig} for complete results
and Sec.~\ref{app:multiclass} for details on the method.

\section{Discussion}
\label{sec:disc}

Most scalable methods for fitting Gaussian process models target the marginal log likelihood.
As we have shown, the fusion of cross-validation and nearest neighbor truncation provides an alternative path to scalability.
The resulting method is simple to implement and offers fast training and excellent predictive performance, both
for regression and classification.
We note three limitations of our approach. First, non-Gaussian likelihoods that do not admit suitable
auxiliary variables may be difficult to accommodate.
Second, as discussed in Sec.~\ref{sec:degenerate}, LOO-$k$ GPs may be inappropriate for highly degenerate datasets.
Third, nearest neighbor queries may be prohibitively slow
for datasets with tens of millions of data points. Modifications to our basic approach may be required to support
this regime, including for example approximate nearest neighbor queries or dataset sharding.


\subsubsection*{Acknowledgements}
GP is supported by the Simons Foundation, McKnight Foundation, the Grossman Center, and the Gatsby Charitable Trust.


\bibliographystyle{plainnat}
\bibliography{loogp}

\appendix
\onecolumn

\section{Appendix}

\begin{algorithm*}
\DontPrintSemicolon 
\KwIn{Dataset $\DD = (\by, \bX)$ with $N$ data points;
      number of nearest neighbors $k$;
      optimizer \texttt{optim};
      mini-batch size $B$;
      number of iterations $T$;
      nearest neighbor index update frequency $T_{\rm nn}$;
      initial kernel hyperparameters $\{\rho_i^0, \sigma_K^0, \sigmaobs^0 \}$
     }
\KwOut{Learned kernel hyperparameters $\{\rho_i^T, \sigma_K^T, \sigmaobs^T \}$ }
\For{$t =1, ..., T$} {
    If $(t-1) \mod T_{\rm nn} = 0$, (re)compute the nearest neighbor index using $\{\rho_i^{t-1}\}$ \\
    Choose a random mini-batch of indices of size $B$: $\II  \subset  \{1, ..., N\}$  with $| \II | = B$ \\
    For each $i \in \II$ form the $k$-nearest-neighbor tuple $\left( \by_{i, k},  \bX_{i, k} \right)$  \\
    Compute a stochastic estimate $\LLLst$ of Eqn.~\ref{eqn:cvk} using Eqn.~\ref{eqn:gppredmean}-\ref{eqn:gppredvar},
                 $\{ \left( \by_{i, k},  \bX_{i, k} \right) \}_{i=1}^B$
             and $\{\rho_i^{t-1}, \sigma_K^{t-1}, \sigmaobs^{t-1} \}$ \\
    Let $\{\rho_i^t, \sigma_K^t, \sigmaobs^t \}$ = \texttt{optim}($\LLLst$) \\
}
    \Return{$\{\rho_i^T, \sigma_K^T, \sigmaobs^T \}$}
    \caption{We outline the main steps in learning a LOO-$k$ GP in the case of univariate regression.}
\label{algo}
\end{algorithm*}

\subsection{Societal impact}
\label{app:impact}

We do not anticipate any negative societal impact from the methods described in this work, although we
note that they inherit the risks that are inherent to any predictive algorithm.
In more detail there is the possibility of the following risks.
First, predictive algorithms can be deployed in ways that disadvantage vulnerable groups in a population.
Even if these effects are unintended, they can still arise if deployed algorithms are poorly vetted with respect to their fairness implications.
Second, algorithms that offer uncertainty quantification may be misused by users who place unwarranted confidence in the
uncertainties produced by the algorithm. This can arise, for example, in the presence of undetected covariate shift.
Third, since our algorithm makes use of non-parametric predictive distributions, it is necessary to retain the training
data to make predictions. This is in contrast to fully parametric models where the training data can be discarded.
As such, nearest neighbor methods may be more vulnerable to data breaches, although standard security practices
should mitigate any such risk.
That said, for these reasons nearest neighbor methods may be inappropriate for use with
sensitive datasets, e.g.~those that contain personally identifiable information.

\subsection{Related work (extended)}
\label{app:related}

Nearest neighbor constructions in the GP context have been explored by several authors.
Vecchia approximations \citep{vecchia1988estimation,katzfuss2021general} exploit a nearest neighbor approximation
to the MLL.
This approach can work well in 2 or 3 dimensions but tends to struggle in higher dimensions due to the
need to choose a fixed ordering of the data.
\citet{datta2016hierarchical} extend the Vecchia approach to the Nearest Neighbor Gaussian Process, which is a valid stochastic process,
deriving a custom Gibbs inference scheme, and illustrating their approach on geospatial data.
\citet{chen2020stochastic} explore a different biased approximation to the MLL, which
does not require specifying an ordering of the data and which we refer to as MLL-$k$. For more details on Vecchia approximations
and MLL-$k$ see the next section, Sec.~\ref{sec:conceptual}.
\citet{tran2021sparse} introduce a variational scheme for GP inference (SWSGP) that leverages nearest neighbor
truncation within an inducing point construction.
Specifically, they introduce a masking latent variable that selects a subset of inducing points for every given data point.
This makes it possible to use many inducing points (unlike conventional inducing point methods like SVGP);
which consequentially introduces many new variational parameters that make optimization more challenging.
\citet{gramacy2015local,gramacy2016lagp}, addressing the computer simulation community, introduce
a `Local Gaussian Process Approximation' for regression that iteratively constructs a nearest neighbor conditioning set
at test time; in contrast to our approach there is no training phase.
Somewhat related to nearest neighbors, several authors have explored the computational
advantages of compactly-supported kernels \citep{melkumyan2009sparse,barber2020sparse}.

Cross-validation (CV) in the GP (or rather kriging) context was explored as early as 1983 by \citet{dubrule1983cross},
with follow-up work including \citet{emery2009kriging}, \citet{zhang2010kriging}, and \citet{le2015cokriging}.
\citet{bachoc2013cross} compares predictive performace of GPs fit with MLL and CV and concludes
that CV is more robust to model mis-specification. The classic GP textbook \citep[][Sec.~5.3]{rasmussen2003gaussian}
also briefly touches on CV in the GP setting.
\citet{smith2016differentially} consider CV losses in the context of differentially private GPs.
Recent work explores how the CV score can be efficiently computed for GP regressors
\citep{ginsbourger2021fast,petit2020towards}.
\citet{vehtari2016bayesian} investigate approximate leave-one-out approaches for gaussian latent variable models.
\citet{jankowiak2020parametric} introduce an inducing point approach for GP regression, PPGPR, that like our approach
uses a loss function that is defined in terms of the predictive distribution. Indeed our approach
can be seen as a non-parametric analog of PPGPR, and Eqn.~\ref{eqn:sigmamll} provides a novel
conceptual framing for that approach.
PPGPR and LOO-$k$ are also related to Direct Loss Minimization, which emerges from a view of
approximate inference as regularized loss minimization \citep{sheth2020pseudo,wei2021direct}.
\citet{fong2020marginal} explore the connection between MLL and CV in the context
of model evaluation and consider a decomposition like that in Eqn.~\ref{eqn:sigmamll}
specialized to the case of exchangeable data. They also advocate
using a Bayesian \emph{cumulative} leave-P-out CV score for fitting models, although the computational
cost limits this approach to small datasets.


Various approaches to approximate GP inference are reviewed in \citet{liu2020gaussian}.
An early application of inducing points is described in \citet{snelson2006sparse},
which motivated various extensions to variational inference \citep{titsias2009variational,hensman2013gaussian,hensman2015scalable}.
\citet{wenzel2019efficient,galy2020multi} introduce an approximate inference scheme
for binary and multi-class classification that exploits inducing points, P\olya-Gamma auxiliary variables, and variational inference.


\subsection{Objective function summary}
\label{sec:conceptual}

\begin{table}[]
\resizebox{\columnwidth}{!}{%
\begin{tabular}{|l|l|l|l|l|l|}
\hline
    {\bf Method} & {\bf Targets MLL?} & {\bf Targets CV?} & {\bf Inducing Points?}  &{\bf Nearest Neighbors?}  &  {\bf Notes} \\ \hline
SVGP    & \Checkmark & \XSolid    & \Checkmark & \XSolid    & Variational lower bound to MLL  \\ \hline
PPGPR   & \XSolid    & \Checkmark & \Checkmark & \XSolid    & CV loss includes additional regularizers \\ \hline
MLL-$k$   & \Checkmark & \XSolid    & \XSolid    & \Checkmark & Does not require fixed dataset ordering \\ \hline
Vecchia & \Checkmark & \XSolid    & \XSolid    & \Checkmark & Requires fixed dataset ordering \\ \hline
SWSGP-$k$ & \Checkmark & \XSolid    & \Checkmark    & \Checkmark & Nearest neighbors w.r.t.~inducing points \\ \hline
ALC & \Checkmark & \XSolid    & \XSolid    & \Checkmark & No training phase \\ \hline
LOO-$k$   & \XSolid    & \Checkmark & \XSolid    & \Checkmark &  \\ \hline
\end{tabular}
    }
    \vspace{2mm}
\caption{We summarize the conceptual differences between seven of the methods for GP regression benchmarked in
    Sec.~\ref{sec:uniexp}.}
    \label{table:conceptual}
\end{table}

In Table~\ref{table:conceptual} we provide a conceptual summary of how the different methods for scalable GP regression
benchmarked in Sec.~\ref{sec:uniexp} are formulated. We note that Vecchia and MLL-$k$ are quite similar, as both utilize
nearest neighbor truncation to target the MLL. The Vecchia approximation does this using a fixed ordering in a decomposition
of the MLL into a product of univariate conditionals, while MLL-$k$ ignores the restrictions that are imposed by a specific
ordering. Both approximations result in biased approximations of the MLL, and for both methods we utilize
mini-batch training. In particular for Vecchia
the objective function is of the form
\begin{equation}
    \LL_{\rm Vecchia} = \frac{1}{N} \sum_{n=1}^N \log p(y_n | \bX_{n, k}^{\rm Vecchia}, \by_{n, k}^{\rm Vecchia}, \bx_n)
\end{equation}
where $\bX_{n, k}^{\rm Vecchia}$ and $\by_{n, k}^{\rm Vecchia}$ are the $k$ nearest neighbors of $\bx_n$
\emph{that respect a given ordering of the data}. For example, for $n^\prime \ge n$ we necessarily
have that $y_{n^\prime} \notin \by_{n, k}$. Since the Vecchia approximation respects a fixed ordering of a data, it
can be understood as a nearest neighbor approximation to a particular decomposition of the MLL.
Conversely for MLL-$k$ the objective function takes the form
\begin{equation}
    \LL_{\rm MLL}^k = \frac{1}{N}  \sum_{n=1}^N \log p(\by_{n, k}^{\rm MLL} | \bX_{n, k}^{\rm MLL})
\end{equation}
where, for example, $ \bX_{n, k}^{\rm MLL}$ consists of the $k-1$ nearest neighbors of $\bx_n$ together with $\bx_n$ itself.

\subsection{A pragmatic view of Bayesian methods}
\label{app:bayesian}

Gaussian processes are often seen from a Bayesian perspective.
This is of course very natural since a Gaussian process makes for a powerful and flexible prior over functions.
However, Gaussian processes can be utilized in a wide variety of applications, and we conceptualize these as occurring along a spectrum.
At one end of the spectrum we can imagine a practitioner we might call the `Bayesian Statistician'. Typically, this statistician
is particularly interested in obtaining high-fidelity posterior approximations. For example, we might imagine constructing a semi-mechanistic
model of air pollution in a city that uses Gaussian processes to model concentrations of different pollutants. Here it might be
of particular interest to compute a posterior probability that a pollutant concentration exceeds a given threshold in a particular area.
On the other end of the spectrum there is a practitioner we might call the `Probabilistic Machine Learner'. This individual is
typically interested in making high quality predictions with well-calibrated uncertainties. Accurate posterior marginals over
various latent variables in the model may be of secondary interest, as the focus is on prediction. We primarily see LOO-$k$ GPs
as being of interest to this second group of practitioners. That is, while one could certainly use a LOO-$k$ approximation
within a typical Bayesian workflow, this choice may be inappropriate for a statistican whose primary concern is with high-fidelity posterior
approximations. For the probabilistic ML practitioner, however, LOO-$k$ GPs have a lot to offer as they represent a simple and robust method
for making well-calibrated predictions on common regression and classification tasks. We expect LOO-$k$ GPs to be particularly useful
in the regime with $N \sim 10^4-10^5$ datapoints, a regime in which there may be too little data to train e.g.~a neural network but too much
data to train a GP using the exact MLL.

\subsection{Computational complexity}
\label{app:complexity}

We provide a brief discussion of the computational complexity of LOO-$k$ as compared to other scalable GP methods.
For simplicity we limit our discussion to the case of univariate regression.
Let $M$ be the number of inducing points, $k$ be the number of nearest neighbors, and $B$ be the size of the mini-batch.
The computational complexity of SVGP and PPGPR are identical. In particular the
time complexity of a training step is $\OO(BM^2 + M^3)$ and the space complexity is $\OO(M^2 + BM)$, while
the time and space complexity at test time (for a single input) are both $\OO(M^2)$.
The computational complexity of Vecchia, MLL-$k$, and LOO-$k$ are identical.
During training, we must compute the set of $k$-nearest neighbors for each training data point, which takes $\OO(N \log N)$ time (or $\OO( N^2)$ time with a brute-force approach).
Once the nearest neighbors are computed, the time complexity of a training step is $\OO(Bk^3)$ and the space complexity is $\OO(Bk^2)$.
During testing, the time and space complexity (for a single input) are $\OO(\log N + k^3)$ and $\OO(k^2)$, respectively.
The $\OO(\log N)$ factor is the time to compute the test input's nearest neighbors ($\OO(N)$ with a brute-force approach), and the $\OO(k^3)$ factor is the cost of computing the posterior mean and variance.


\subsection{P\olya-Gamma density}
\label{app:pgdensity}

The probability density function of the P\olya-Gamma distribution $p(\omega|1,0)$, which has support
on the positive real line, is given
by the alternating series \citep{polson2013bayesian}:
\begin{equation}
    p(\omega|1,0) = \sum_{n=0}^\infty (-1)^n \frac{2 n + 1}{\sqrt{2\pi \omega^3}}
    \exp\left(-(2n+1)^2/8\omega\right)
    \label{eqn:pgdensity}
\end{equation}
The mean of this distribution is given by $\tfrac{1}{4}$ and the vast majority of the probability mass
is located in the interval $\omega \in (0, 2.5)$. To perform variational inference w.r.t.~$p(\omega|1,0)$
we need to be able to compute this density.
We leverage the implementation in Pyro \citep{bingham2019pyro}, which is formulated as follows.
Estimating Eqn.~\ref{eqn:pgdensity} accurately requires computing increasingly many terms
in the alternating sum as $\omega$ increases. To address this issue we truncate the distribution to the interval
$(0, 2.5)$. We then retain the leading 7 terms in Eqn.~\ref{eqn:pgdensity}. This is accurate to about 6 decimal
places over the entire truncated domain. We find that this approximation is sufficient for our purposes. We
note that as a consequence of the truncation our variational distribution is properly speaking a truncated log-Normal distribution,
although given the large truncation point and given that most of the posterior mass concentrates in $\omega \in (0, 0.5)$
the truncation of $q(\bomega)$ plays a negligible role numerically and can be safely ignored.

\subsection{Binary classification}
\label{app:binclass}

We expand on our method for binary GP classification discussed in Sec.~\ref{sec:class}.
We consider a dataset $\DD = \{(\bx_n, y_n)\}_{n=1}^N$ with $y_n \in \{-1, 1\}$ and a GP with joint density given by
\begin{equation}
 p(\bff | \bK_{N,N}) \prod_{n=1}^N p(y_n | f(\bx_n))
\end{equation}
 where $p(\bff | \bK_{N,N})$ is the GP prior,
 and $p(y_n | f(\bx_n)) = \left[1 + \exp(-y_n f(\bx_n))\right]^{-1}$ is a Bernoulli probability governed by a logistic link function.
 We introduce a $N$-dimensional vector of P\olya-Gamma auxiliary variables $\bomega$ and exploit the identity  \citep{polson2013bayesian}
\begin{equation}
 \frac{e^\psi}{1 + e^\psi} = \tfrac{1}{2} e^{\tfrac{1}{2}\psi} \EE_{p(\omega | 1, 0)} \left[ e^{-\tfrac{1}{2} \omega \psi^2} \right]
\end{equation}
  and the shorthand $f_n = f(\bx_n)$ to write
\begin{align}
    &\prod_{n=1}^N p(y_n | f_n) =
    \prod_{n=1}^N \frac{1}{1 + \exp(-y_n f_n)}
    =\prod_{n=1}^N \frac{\exp(y_n f_n)}{1 + \exp(y_n f_n)} \\
    &= 2^{-N} \prod_{n=1}^N \EE_{p(\omega_n | 1, 0)} \left[
                         \exp(\tfrac{1}{2} y_n f_n - \tfrac{1}{2} \omega_n y_n^2 f_n^2) \right] \\
    & \propto \prod_{n=1}^N \EE_{p(\omega_n | 1, 0)} \left[
                         \exp(\tfrac{1}{2} y_n f_n - \tfrac{1}{2} \omega_n f_n^2) \right]
    \propto \EE_{p(\bomega)} \left[  \prod_{n=1}^N
             \exp\left(\tfrac{1}{2} \by^{T} \bff - \tfrac{1}{2}  \bff^{T} \bOmega \bff  \right) \right]
\end{align}
 where $\bOmega \equiv {\rm diag}(\bomega)$ is a diagonal $N \times N$ matrix.
 Crucially, thanks to the P\olya-Gamma augmentation $\bff$ is now conditionally gaussian when
 we condition on $\by$ and $\bomega$. We emphasize that this augmentation is exact.
 Next we integrate out $\bff$. This is made easy if we recycle familiar formulae from the GP regression case.
 In particular write
\begin{equation}
    \tfrac{1}{2} \by^{T} \bff - \tfrac{1}{2}  \bff^{T} \bOmega \bff
    = -\tfrac{1}{2} \sum_n \omega_n \left(f_n - \frac{y_n}{2 \omega_n}\right)^2  + \sum_n \frac{1}{8 \omega_n}
\end{equation}
and observe that (apart from the last term which is independent of $\bff$)
this takes the form of a Normal likelihood with `pseudo-observations' $\frac{y_n}{2 \omega_n}$ and
data point dependent observation variances $\omega_n^{-1}$.
From this we can immediately write down the marginal log likelihood:
\begin{align}
    \log p(\by | \bX) = \log \EE_{p(\bomega)} p(\by | \bX, \bomega) \qquad {\rm with} \\
    p(\by | \bX, \bomega) = \NN \left(\frac{1}{2} \bOmega^{-1}\by, \bK_{N,N}  + \bOmega^{-1} \right)
    \times  \exp \left(\Sigma_n \tfrac{1}{8 \omega_n} \right)
\end{align}
Next we introduce a variational distribution $q(\bomega)$ and apply Jensen's inequality to obtain
\begin{align}
    \log    p(\by | \bX) \ge \EE_{q(\bomega)} \left[ \log p(\by | \bX, \bomega) \right] - {\rm KL}(q(\bomega) | p(\bomega) )
    + \tfrac{1}{8} \EE_{q(\bomega)} \left[ \Sigma_n \omega_n^{-1}\right]
\end{align}
where we choose $q(\bomega)$ to be a (truncated) mean-field log-Normal distribution (see the discussion
in Sec.~\ref{app:pgdensity}).
We note that if $q(\omega)$ is parameterized with location and scale parameters $m$ and $s$ then
$\EE_{q(\omega)} \left[\omega^{-1}\right]=\exp(-m+\tfrac{1}{2}s^2)$.
This expectation is unbounded from above as $s \rightarrow \infty$ or $m \rightarrow -\infty$, corresponding
to putting lots of posterior mass near $\omega=0$. This is potentially an issue for our variational procedure,
since it can potentially lead to undesired run-away solutions. One way to address this isue is to limit
ourselves to variational distributions that are sufficiently well-behaved at $\omega=0$. For example,
we could truncate the log-Normal distribution at some finite $\omega_{\rm min}$. Here we take a simpler
approach and omit the $\EE_{q(\omega)} \left[\omega^{-1}\right]$ term from our variational objective,
noting that this term is always positive so that our modified variational objective is still guaranteed
to be a lower bound to the log evidence. In practice we find that this approach works well. A different
approach would lead to slightly different regularization of $\omega$ but we do not expect this to be an important
effect. Indeed what's most important here is that $\omega$ allows us to target a closed-form cross-validation-based objective
and formulate non-parametric predictive distributions.
Next we replace $\log p(\by | \bX, \bomega)$ with its leave-one-out approximation to obtain
an objective
\begin{align}
\LLL = \frac{1}{N} \sum_{n=1}^N \EE_{q(\bomega)} \left[  \log p(y_n | \by_{-n}, \bX, \bomega_{-n}) \right]  - {\rm KL}(q(\bomega) | p(\bomega) )
    \label{eqn:llomega}
\end{align}
where the (conditional) posterior predictive distribution $p(y_n | \by_{-n}, \bX, \bomega_{-n})$ is given by
 \begin{equation}
p(y_n | \by_{-n}, \bX, \bomega_{-n}) \equiv \int \! df_n \; p(y_n | f_n) p(f_n | \by_{-n}, \bX, \bomega_{-n})
\label{eqn:postpredynapp}
\end{equation}
 and where the posterior over the latent function value $f_n$, namely
$p(f_n | \by_{-n}, \bX, \bomega_{-n})$, is given by the Normal distribution
$\NN(f_n  | \mu_n, \sigma_n^2)$
with mean and variance equal to
\begin{align}
\mu_n &= \frac{1}{2} \bK_{-n, n}^{\rm T} \left( \bK_{-n, -n}+ \bOmega_{-n}^{-1} \right)^{-1} \bOmega_{-n}^{-1} \by_{-n} \\
\sigma_n^2 &=  K_{n,n}-\bK_{-n, n}^{\rm T} \left( \bK_{-n,-n}+ \bOmega_{-n}^{-1} \right)^{-1} \bK_{-n, n}
\end{align}
where $\bOmega_{-n}$ is the $(N-1)\times(N-1)$ diagonal matrix that omits $\omega_n$ and
\begin{align}
    \bK_{-n,-n} &\equiv K(\bX_{-n}, \bX_{-n}) \in \RR^{n-1,n-1} \\
    \bK_{-n, n} &\equiv K(\bX_{-n}, x_n)  \in \RR^{n-1,1} \qquad K_{n,n} \equiv K(x_n, x_n)
    \label{eqn:kdefs}
\end{align}
To obtain our final objective function we apply a $k$-nearest-neighbor truncation to
Eqn.~\ref{eqn:llomega}-\ref{eqn:kdefs}. For example this means that for each $n$
the expression for the predictive distribution
$p(y_n | \by_{-n}, \bX, \bomega_{-n})$ will be replaced by an expression that only depends
on the $k$ P\olya-Gamma variates that correspond to its nearest neighbors.
We numerically approximate the univariate integral in Eqn.~\ref{eqn:postpredynapp} with Gauss-Hermite quadrature using
$Q$ quadrature points.

To maximize this objective function we use standard techniques from stochastic variational inference, including data subsampling
 and reparameterized gradients of $\bomega$.
 At test time predictions can be obtained by computing Eqn.~\ref{eqn:postpredynapp} after
 conditioning on a sample $\bomega \sim q(\bomega)$.
 In practice we use a single sample, since we found negligible gains from averaging over multiple samples.
 The computational cost of a training iteration is $\OO(Bk^3 + BQ)$, where $B$ is the mini-batch size.
 Note that (excluding the cost of finding the nearest neighbors)
 the cost at test time for a batch of test points of size $B$ is also $\OO(Bk^3 + BQ)$,
 since the main cost of computing the objective function goes into computing the predictive distribution.
 In practice we take $Q=16$ so that the cost of Gauss-Hermite quadrature is negligible.
 Consequently the main determinants of computational cost are $B$ and $k$.

\subsection{Multi-class classification}
\label{app:multiclass}

We are given a dataset $\DD = \{ (\bx_n, \by_n) \}$, where each $\by_n$ encodes one of $K$ discrete
labels and we suppose that $\by_n$ is represented as a one-against-all encoding
where $\by_n \in \{ -1, 1 \}^K$ and $y_{n,k} = 1$ for exactly one of $k \in \{1, ..., K\}$.
In the following we will effectively learn $K$ one-against-all GP classifiers constructed as in Sec.~\ref{app:binclass},
with the difference that we will aggregate the $K$ one-against-all classification probabilities and form a single
multi-class likelihood.

In more detail we introduce a $N \times K$ matrix of P\olya-Gamma auxiliary variables $\bomega_{n,k}$.
As in Eqn.~\ref{eqn:postpredyn} for each data point $n$ we can form $K$ Bernoulli distributions
 \begin{equation}
     \tilde{p}(y_{n,k} | \by_{-n, k}, \bX, \bomega_{-n,k}) \equiv \int \! df_{n,k} \; p(y_{n,k} | f_{n,k})
                                                            p(f_{n,k} | \by_{-n,k}, \bX, \bomega_{-n,k })
\label{eqn:postpredynk}
\end{equation}
where each of the $K$ univariate integrals can be numerically approximated with Gauss-Hermite quadrature.
We use a tilde in Eqn.~\ref{eqn:postpredynk} to indicate that $\tilde{p}(y_{n,k} | \cdot)$ is an intermediate quantity
that serves as an ingredient in computing a joint predictive distribution. In particular we form a joint predictive
distribution by jointly normalizing all of the Bernoulli probabilities:
\begin{equation}
     p(y_{n, k} = 1 | \by_{-n}, \bX, \bomega_{-n} ) = \frac{ \tilde{p}(y_{n,k} = 1 | \cdot) }{ \sum_{k^\prime} \tilde{p}(y_{n,k^\prime} = 1 | \cdot) }
\label{eqn:postpredynknorm}
\end{equation}
Given the matrix of P\olya-Gamma variates $\bomega_{n,k}$, $p(y_{n, k} = 1 | \cdot)$
in Eqn.~\ref{eqn:postpredynknorm} is a normalized distribution over the label $k \in \{1, ..., K \}$ that can be computed
in closed form thanks to Gauss-Hermite quadrature. To train the kernel hyperparameters and P\olya-Gamma mean-field variational distribution
we use an objective function that is a direct generalization of Eqn.~\ref{eqn:loobin}:
 \begin{equation}
     \LLL^{\rm multi} \equiv \frac{1}{N} \sum_{n=1}^N \EE_{q(\bomega)} \left[  \log p(y_{n, k} | \by_{-n}, \bX, \bomega_{-n}) \right] - {\rm KL}(q(\bomega) | p(\bomega) )
\label{eqn:loomulti}
\end{equation}
The final objective function is obtained by applying a $k$-nearest-neighbors truncation to Eqn.~\ref{eqn:loomulti}.
During test time we sample $\bomega \sim q(\bomega)$ and use Eqn.~\ref{eqn:postpredynknorm}.

\subsection{Other likelihoods}
\label{app:other}

The auxiliary variable construction in Sec.~\ref{sec:class}-\ref{sec:multiclass} can be extended to a number of
other likelihoods. For example, P\olya-Gamma auxiliary variables can also be used to accomodate
binomial and negative binomial likelihoods \citep{polson2013bayesian}.
Additionally, gamma auxiliary variables can be used to accomodate
a Student's t likelihood, which is useful for modeling heavy-tailed noise.

\subsection{Experimental details}
\label{app:exp}

All the datasets we use, apart from those used in the multivariate regression experiments, can be obtained
from the UCI depository \citep{Dua:2019}.
The Fw.-Kuka, Fw.-Baxter, and Rythmic-Baxter multivariate regression datasets are available from \url{https://bitbucket.org/athapoly/datasets/src/master/},
and the Sarcos dataset is available from \url{http://www.gaussianprocess.org/gpml/data/}.

\subsubsection{General training details}
\label{app:exptrain}

For all experiments the GP model we fit
uses a Mat{\'e}rn 5/2 kernel with individual length scales for each input dimension.
For all regression experiments the prior GP mean is a learnable constant; otherwise it is fixed to zero.
For all experiments we use the Adam optimizer \citep{kingma2014adam}. When
training a GP with MLL we set Adam's momentum hyperparameter $\beta_1$ to
$\beta_1 = 0.5$; otherwise we set $\beta_1 = 0.90$. We use a stepwise learning rate schedule
in which the learning rate starts high ($0.03$) and is reduced by a factor of 5 after $25\%$, $50\%$, and $75\%$ of optimization.
Our default batch size is $B=128$, although we use $B=64$ when
the computational demands are higher (e.g.~for multi-class classification).
We use FAISS for nearest neighbor queries \citep{johnson2019billion}.
During training we update nearest neighbor indices every $50$ gradient steps, although
we note that a smaller update frequency also works well.

To run our experiments we used a small number of GPUs, including a GeForce RTX 2080 GPU, a Quadro RTX 5000, and
a GeForce GTX 1080 Ti. We estimate that we used $\sim\! 200$ GPU hours running pilot and final experiments.

\subsubsection{Details for particular experiments}
\label{app:expdetails}

\paragraph{Model mis-specification}
\label{app:expmis}

The GP prior we use to generate synthetic data has length scale $\rho = \thalf$.
Note we generate a single data set and then apply different warpings to it; i.e.~results
for different $\gamma$ differ only in the warping applied to the inputs.

\paragraph{Objective function comparison}
\label{app:expobj}

Inducing point locations are set using k-means clustering.
We retrain SVGP variational parameters for each hyperparameter setting, while keeping the $M=100$
inducing point locations fixed.

\paragraph{Dependence on number of nearest neighbors $k$}
\label{app:expk}

Fig.~\ref{fig:k} uses the results reported in Sec.~\ref{sec:uniexp}, with the difference
that additional runs with smaller/larger $k$ are included.

\paragraph{Runtime performance}

To define a regression task we use the Slice UCI dataset and downsample the input
dimension to $D=25$. To accommodate large $N$
we also enlarge the dataset by repeating individual data points and adding noise
(both to inputs and targets).
We use a mini-batch size of $B=128$ and run on a GeForce RTX 2080 GPU.
As in our other experiments, we update the nearest neighbor index every 50 gradient steps.

\paragraph{Degenerate data regime}

For each (univariate regression) dataset we duplicate each data point in the training set and add zero mean
gaussian noise (with variance given by $10^{-4}$) to each input $\bx_n$ and each response $y_n$.
Other details are as in the next section.

\paragraph{Univariate regression}

Since we reproduce some of the baseline results from \citep{jankowiak2020parametric}, we follow
the experimental procedure detailed there.
In particular regression datasets are centered and normalized so that the trivial zero prediction has a mean squared error of unity.
All the experiments in Sec.~\ref{sec:uniexp}-\ref{sec:multiexp} use training/test/validation
splits with proportions 15:3:2.
For LOO-$k$ we vary $k \in \{32, 64, 128, 256 \}$ and use the validation set LL to choose the best $k$.
SVGP uses $M=1000$ inducing points with $\beta_{\rm reg} \in \{0.1, 0.3, 0.5, 1.0 \}$ where
 $\beta_{\rm reg}$ is a scaling term in front of the KL divergence.
PPGPR (specifically the MFD variant, see \citep{jankowiak2020parametric}) uses $M=1000$ inducing points with $\beta_{\rm reg} \in \{0.01, 0.05, 0.2, 1.0 \}$.
Both SVGP and PPGPR results are reproduced from \citep{jankowiak2020parametric}.
In both cases inducing point locations are initialized with $k$-means.
For Vecchia and MLL-$k$ we also vary $k \in \{32, 64, 128, 256 \}$.
For the Vecchia baseline, the data are ordered according to the first PCA vector.
For SWSGP, we use $M=10000$ inducing points (a order of magnitude more than SVGP) since this method scales linearly with $M$.
Following \citet{tran2021sparse}, we parameterize the variational distribution $q(u_1, \ldots, u_M)$ to be independent Gaussians.
We vary $k \in \{32, 64, 128, 256 \}$, and we jointly optimize the variational parameters and hyperparameters for $6400$ iterations.
For ALC we use the software described in \citep{gramacy2016lagp}. Since prediction is slow, both because the implementation
is CPU only and because the iterative procedure is inherently expensive,
we use a fixed number of $k=64$ nearest neighbors.
For ALC we use an istropic kernel on Kegg-undirected because we were unable to obtain reasonable results with a kernel
with per-dimension length scales on this particular dataset.

\paragraph{Multivariate regression}

Since we reproduce some of the baseline results from \citep{jankowiak2020deep}, we follow
the experimental procedure detailed there.
In particular regression datasets are centered and normalized so that the trivial zero prediction has
a mean RMSE of unity (i.e.~the RMSE averaged across all output dimensions).
For LOO-$k$ we vary $k \in \{16, 32 \}$ and use the validation set LL to choose the best $k$.
For MLL-$k$ we vary $k \in \{16, 32 \}$.
SVGP and PPGPR both use $M=300$ inducing points with $\beta_{\rm reg} \in \{0.1, 0.3, 0.5, 1.0 \}$.

\paragraph{Binary classification}

For LOO-$k$ we vary $k \in \{32, 64, 128, 256 \}$ and use the validation set LL to choose the best $k$.
For the SVGP and the P\olya-Gamma PGVI baseline we use $M=1024$ inducing points and vary $\beta_{\rm reg} \in \{0.1, 0.3, 0.5, 1.0 \}$.
We subsample the SUSY and Higgs datasets down to $N=2 \times 10^5$ data points for simplicity.
The Covtype dataset is inherently a multi-class dataset with $K=7$ datasets. We convert it into two binary classification
datasets by combining the two Pine and Fir classes, respectively, in a two-against-five fashion to obtain
two derived datasets, Covtype-Pine and Covtype-Fir, respectively.
We use $Q=16$ Gauss-Hermite quadrature points.

\paragraph{Multi-class classification}

For LOO-$k$ we vary $k \in \{32, 64, 128, 256 \}$ and use the validation set LL to choose the best $k$.
For SVGP we use $M=512$ inducing points and vary $\beta_{\rm reg} \in \{0.1, 0.3, 0.5, 1.0 \}$.
We use $Q=16$ Gauss-Hermite quadrature points.

\subsection{Additional figures}
\label{app:addfig}

\begin{figure}
  \centering
  \includegraphics[width=0.85\linewidth]{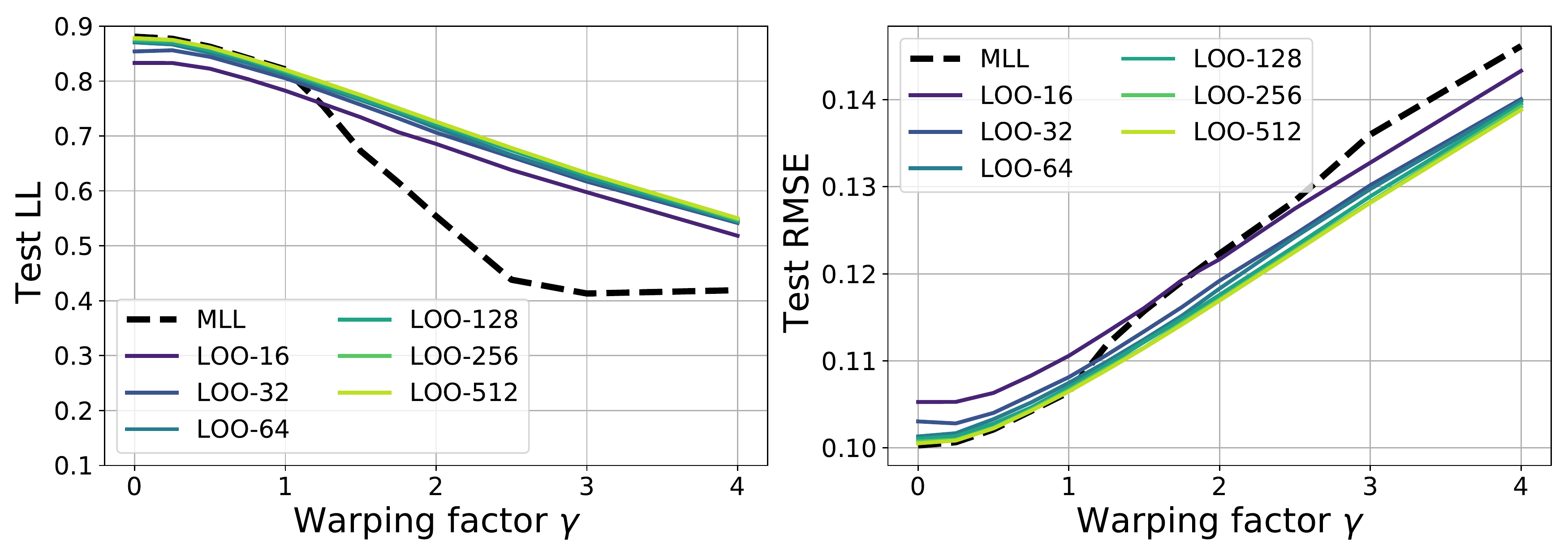}
  \includegraphics[width=0.85\linewidth]{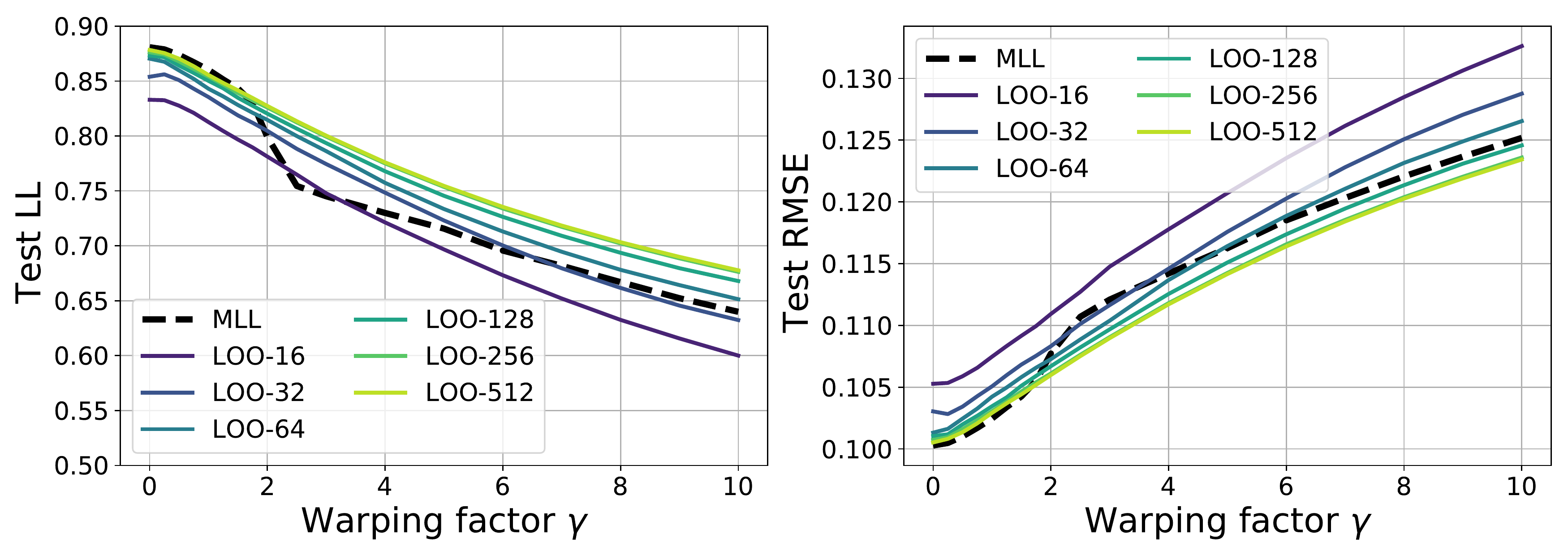}
        \caption{These two figures are companions to Fig.~\ref{fig:mis} in the main text.
        We compare predictive performance of a GP trained with a LOO-$k$ objective to a GP
    trained via MLL, where the GP regressor is mis-specified due
  to non-stationarity in the dataset introduced by a warping function controlled by $\gamma$.
    {\bf Top}: The coordinatewise warping function is the identity for $x_i \le 0$ and
    $x_i \to x_i^{1 + \gamma}$ otherwise.
    {\bf Bottom}: The coordinatewise warping function is $x_i \to x_i + \gamma x_i^3$.
    In both cases as $\gamma$ increases and the dataset becomes more non-stationary,
    the LOO-$k$ GP exhibits superior predictive performance, both w.r.t.~log likelihood (LL) and root mean squared error (RMSE).}
    \label{fig:appwarped}
\end{figure}

\begin{figure}
  \centering
  \includegraphics[width=\linewidth]{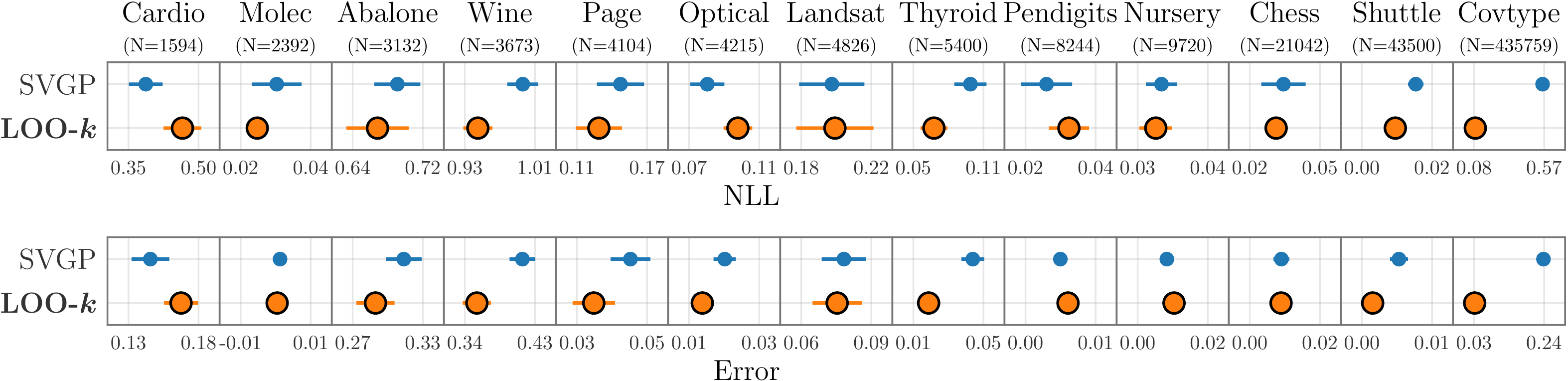}\\
  \caption{We depict predictive negative log likelihoods (NLL, top) and
    predictive error (bottom) for 13 multi-class classification datasets
    using LOO-$k$ and SVGP.
    Results are averaged over 5 dataset splits.
    Lower numbers are better.}
  \label{fig:multi_classification}
\end{figure}

In Fig.~\ref{fig:appwarped} we replicate the experiment described in
Sec.~\ref{sec:misexp} for two additional warping functions.
In Fig.~\ref{fig:multi_classification} we depict the results for the multi-class classification
experiment in Sec.~\ref{sec:multiexp}.

\begin{figure}
\floatbox[{\capbeside\thisfloatsetup{capbesideposition={left,top},capbesidewidth=7.5cm}}]{table}[\FBwidth]
{\caption{
  We summarize the performance ranking of LOO-$k$ and SVGP on the multi-class classification datasets in Figure~\ref{fig:multi_classification},
  averaged over all splits.
}\label{tab:multi_classification_rank}}
{ 
  \vspace{-1.5mm}
  \resizebox{0.70\linewidth}{!}{%
    \begin{tabular}{cccc}
\toprule
{} &   &    SVGP &            LOO-$k$ \\
Value &           &         &                    \\
\midrule
NLL   &   &  $1.63$ &  $\mathbf{ 1.37 }$ \\
Error &   &  $1.68$ &  $\mathbf{ 1.32 }$ \\
\bottomrule
\end{tabular}

  }
} 
\end{figure}

\subsection{Results tables}
\label{app:tables}

Tables~\ref{tab:uci_regression_full}, \ref{tab:multi_regression_full}, and \ref{tab:classification_full}
report results for the principal regression and classification experiments from Section~\ref{sec:exp}.
See Table~\ref{table:degenerate} for complete results for the experiment in Sec.~\ref{sec:degenerate}.

\begin{table}
  \centering
  \caption{
      \hspace{-1.10em}
    We summarize the results for the experiment in Sec.~\ref{sec:degenerate} exploring
    the performance of LOO-64 and SVGP-512 on degenerate datasets. We report both test log likelihood and
    RMSE on held out test data.}
    \label{table:degnerate}
  \resizebox{0.8\linewidth}{!}{%
\begin{tabular}{cc|cc|ccccc}
\toprule
    {} &   &    {\bf Original} &      & {\bf Degenerate} &       &  \\
    {\bf Dataset} & {\bf Method}  &    Log Likelihood   & RMSE & Log Likelihood     &  RMSE &  \\
\midrule
Bike   & SVGP-512   &  $0.859$     &  $0.082$ &  $0.553$ &  $0.118$ &    \\
Bike  & LOO-64 &  $2.498$     &  $0.50$ &  $-0.912$ &  $0.224$ &  \\ 
Elevators   & SVGP-512   &  $-1.111$     &  $0.505$ &  $-1.107$ &  $0.493$ &    \\
Elevators  & LOO-64 &  $-0.414$     &  $0.369$ &  $-3.783$ &  $0.676$ &  \\ 
Pol   & SVGP-512   &  $-1.040$     &  $0.367$ &  $-1.054$ &  $0.393$ &    \\
Pol  & LOO-64 &  $1.162$     &  $0.077$ &  $0.689$ &  $0.128$ &  \\ 
Slice   & SVGP-512   &  $0.611$     &  $0.110$ &  $0.726$ &  $0.099$ &    \\
Slice& LOO-64 &  $2.723$     &  $0.023$ &  $2.787$ &  $0.023$ &  \\ 
\bottomrule
\end{tabular}
  }
    \label{table:degenerate}
\end{table}


\begin{table}
  \centering
  \caption{
    A compilation of all univariate regression results from Section~\ref{sec:uniexp}.
    Numbers are averages $\pm$ standard errors over dataset splits.
  }
  \label{tab:uci_regression_full}
  \resizebox{\linewidth}{!}{%
    \begin{tabular}{cccccccccccc}
\toprule
     &            &   &                           MLL &                          SVGP &                          PPGPR &                       Vecchia &               SWSGP &                            ALC &                       MLL-$k$ &                        LOO-$k$ \\
Metric & Dataset &           &                               &                               &                                &                               &                     &                                &                               &                                \\
\midrule
\midrule
NLL & Pol &   &            $-0.817 \pm 0.001$ &            $-0.651 \pm 0.005$ &             $-0.825 \pm 0.005$ &            $-1.190 \pm 0.015$ &  $-1.036 \pm 0.022$ &  $\mathbf{ -1.696 \pm 0.012 }$ &            $-1.200 \pm 0.014$ &             $-1.238 \pm 0.018$ \\
     & Elevators &   &             $0.398 \pm 0.013$ &             $0.401 \pm 0.007$ &   $\mathbf{ 0.329 \pm 0.007 }$ &             $0.407 \pm 0.007$ &   $0.651 \pm 0.015$ &              $0.441 \pm 0.012$ &             $0.397 \pm 0.007$ &              $0.401 \pm 0.008$ \\
     & Bike &   &            $-1.750 \pm 0.014$ &            $-0.807 \pm 0.007$ &             $-1.314 \pm 0.007$ &            $-1.743 \pm 0.053$ &  $-1.401 \pm 0.020$ &             $-1.397 \pm 0.013$ &            $-2.541 \pm 0.008$ &  $\mathbf{ -2.771 \pm 0.031 }$ \\
     & Kin40K &   &            $-0.075 \pm 0.001$ &            $-0.414 \pm 0.002$ &             $-0.770 \pm 0.002$ &            $-0.763 \pm 0.002$ &  $-0.694 \pm 0.008$ &             $-0.778 \pm 0.003$ &            $-0.884 \pm 0.002$ &  $\mathbf{ -1.040 \pm 0.030 }$ \\
     & Protein &   &             $0.875 \pm 0.002$ &             $0.902 \pm 0.003$ &              $0.747 \pm 0.008$ &             $0.936 \pm 0.011$ &   $0.804 \pm 0.006$ &              $0.826 \pm 0.020$ &  $\mathbf{ 0.635 \pm 0.007 }$ &   $\mathbf{ 0.626 \pm 0.008 }$ \\
     & Keggdir. &   &            $-0.870 \pm 0.052$ &            $-1.045 \pm 0.017$ &  $\mathbf{ -1.601 \pm 0.016 }$ &            $-1.048 \pm 0.016$ &  $-1.033 \pm 0.019$ &                            --- &            $-1.055 \pm 0.019$ &             $-1.068 \pm 0.015$ \\
     & Slice &   &            $-1.240 \pm 0.004$ &            $-1.267 \pm 0.003$ &             $-1.516 \pm 0.004$ &            $-1.648 \pm 0.007$ &  $-1.797 \pm 0.016$ &             $-2.441 \pm 0.003$ &            $-2.572 \pm 0.002$ &  $\mathbf{ -2.753 \pm 0.027 }$ \\
     & Keggundir. &   &            $-0.643 \pm 0.008$ &            $-0.704 \pm 0.006$ &  $\mathbf{ -1.807 \pm 0.018 }$ &            $-0.696 \pm 0.004$ &  $-0.690 \pm 0.005$ &              $0.498 \pm 0.310$ &            $-0.697 \pm 0.005$ &             $-0.714 \pm 0.006$ \\
\midrule
RMSE & Pol &   &  $\mathbf{ 0.074 \pm 0.001 }$ &             $0.107 \pm 0.002$ &              $0.111 \pm 0.002$ &  $\mathbf{ 0.075 \pm 0.002 }$ &   $0.091 \pm 0.002$ &              $0.083 \pm 0.001$ &             $0.077 \pm 0.002$ &   $\mathbf{ 0.073 \pm 0.001 }$ \\
     & Elevators &   &  $\mathbf{ 0.347 \pm 0.007 }$ &             $0.360 \pm 0.003$ &              $0.362 \pm 0.003$ &             $0.362 \pm 0.003$ &   $0.448 \pm 0.004$ &              $0.376 \pm 0.003$ &             $0.359 \pm 0.003$ &              $0.360 \pm 0.003$ \\
     & Bike &   &  $\mathbf{ 0.022 \pm 0.001 }$ &             $0.088 \pm 0.002$ &              $0.094 \pm 0.002$ &             $0.029 \pm 0.002$ &   $0.059 \pm 0.002$ &              $0.098 \pm 0.002$ &  $\mathbf{ 0.025 \pm 0.002 }$ &              $0.028 \pm 0.002$ \\
     & Kin40K &   &  $\mathbf{ 0.084 \pm 0.005 }$ &             $0.147 \pm 0.001$ &              $0.188 \pm 0.002$ &             $0.117 \pm 0.000$ &   $0.121 \pm 0.001$ &              $0.129 \pm 0.001$ &             $0.105 \pm 0.000$ &              $0.095 \pm 0.003$ \\
     & Protein &   &  $\mathbf{ 0.514 \pm 0.003 }$ &             $0.594 \pm 0.002$ &              $0.609 \pm 0.002$ &             $0.595 \pm 0.007$ &   $0.542 \pm 0.003$ &              $0.571 \pm 0.002$ &             $0.525 \pm 0.002$ &              $0.526 \pm 0.003$ \\
     & Keggdir. &   &             $0.090 \pm 0.004$ &  $\mathbf{ 0.086 \pm 0.001 }$ &              $0.089 \pm 0.002$ &  $\mathbf{ 0.085 \pm 0.001 }$ &   $0.087 \pm 0.001$ &                            --- &  $\mathbf{ 0.085 \pm 0.001 }$ &   $\mathbf{ 0.084 \pm 0.001 }$ \\
     & Slice &   &             $0.031 \pm 0.003$ &             $0.051 \pm 0.001$ &              $0.095 \pm 0.001$ &             $0.038 \pm 0.002$ &   $0.041 \pm 0.001$ &              $0.026 \pm 0.002$ &  $\mathbf{ 0.021 \pm 0.001 }$ &              $0.024 \pm 0.001$ \\
     & Keggundir. &   &  $\mathbf{ 0.119 \pm 0.002 }$ &             $0.120 \pm 0.001$ &              $0.124 \pm 0.001$ &             $0.120 \pm 0.001$ &   $0.124 \pm 0.001$ &   $\mathbf{ 0.118 \pm 0.001 }$ &             $0.121 \pm 0.001$ &   $\mathbf{ 0.117 \pm 0.001 }$ \\
\midrule
CRPS & Pol &   &             $0.051 \pm 0.000$ &             $0.061 \pm 0.000$ &              $0.057 \pm 0.000$ &             $0.039 \pm 0.000$ &   $0.042 \pm 0.001$ &   $\mathbf{ 0.034 \pm 0.000 }$ &             $0.039 \pm 0.000$ &              $0.036 \pm 0.001$ \\
     & Elevators &   &  $\mathbf{ 0.195 \pm 0.003 }$ &             $0.198 \pm 0.001$ &   $\mathbf{ 0.193 \pm 0.001 }$ &             $0.199 \pm 0.001$ &   $0.244 \pm 0.002$ &              $0.205 \pm 0.002$ &             $0.197 \pm 0.001$ &              $0.197 \pm 0.001$ \\
     & Bike &   &             $0.019 \pm 0.000$ &             $0.049 \pm 0.000$ &              $0.037 \pm 0.000$ &             $0.018 \pm 0.001$ &   $0.026 \pm 0.000$ &              $0.040 \pm 0.000$ &             $0.009 \pm 0.000$ &   $\mathbf{ 0.008 \pm 0.000 }$ \\
     & Kin40K &   &             $0.093 \pm 0.000$ &             $0.082 \pm 0.000$ &              $0.077 \pm 0.000$ &             $0.063 \pm 0.000$ &   $0.063 \pm 0.000$ &              $0.066 \pm 0.000$ &             $0.054 \pm 0.000$ &   $\mathbf{ 0.047 \pm 0.001 }$ \\
     & Protein &   &             $0.293 \pm 0.001$ &             $0.326 \pm 0.001$ &              $0.310 \pm 0.001$ &             $0.333 \pm 0.004$ &   $0.285 \pm 0.001$ &              $0.285 \pm 0.001$ &  $\mathbf{ 0.259 \pm 0.001 }$ &   $\mathbf{ 0.260 \pm 0.001 }$ \\
     & Keggdir. &   &             $0.046 \pm 0.002$ &             $0.037 \pm 0.000$ &   $\mathbf{ 0.031 \pm 0.000 }$ &             $0.038 \pm 0.000$ &   $0.038 \pm 0.000$ &                            --- &             $0.037 \pm 0.001$ &              $0.035 \pm 0.000$ \\
     & Slice &   &             $0.029 \pm 0.000$ &             $0.031 \pm 0.000$ &              $0.032 \pm 0.000$ &             $0.022 \pm 0.000$ &   $0.019 \pm 0.000$ &              $0.012 \pm 0.000$ &             $0.009 \pm 0.000$ &   $\mathbf{ 0.009 \pm 0.000 }$ \\
     & Keggundir. &   &             $0.056 \pm 0.001$ &             $0.051 \pm 0.000$ &              $0.036 \pm 0.000$ &             $0.053 \pm 0.000$ &   $0.052 \pm 0.000$ &   $\mathbf{ 0.033 \pm 0.000 }$ &             $0.051 \pm 0.000$ &              $0.050 \pm 0.000$ \\
\bottomrule
\end{tabular}

  }
\end{table}

\begin{table}
  \centering
  \caption{
    A compilation of all multivariate regression results from Section~\ref{sec:multivarexp}.
    Numbers are averages $\pm$ standard errors over dataset splits.
  }
  \label{tab:multi_regression_full}
  \resizebox{0.9\linewidth}{!}{%
    \begin{tabular}{cccccccc}
\toprule
     &        &   &                          SVGP &               PPGPR &                        MLL-$k$ &                        LOO-$k$ \\
Metric & Dataset &           &                               &                     &                                &                                \\
\midrule
\midrule
NLL & Rt.-Baxter &   &            $-0.003 \pm 0.008$ &  $-0.087 \pm 0.001$ &  $\mathbf{ -1.123 \pm 0.004 }$ &             $-0.911 \pm 0.006$ \\
     & Fw.-Baxter &   &            $-1.353 \pm 0.001$ &  $-1.717 \pm 0.001$ &             $-1.936 \pm 0.001$ &  $\mathbf{ -1.961 \pm 0.000 }$ \\
     & Fw.-Kuka &   &            $-1.060 \pm 0.001$ &  $-1.557 \pm 0.001$ &  $\mathbf{ -1.892 \pm 0.003 }$ &  $\mathbf{ -1.886 \pm 0.003 }$ \\
     & Sarcos &   &            $-0.862 \pm 0.000$ &  $-1.095 \pm 0.001$ &             $-1.439 \pm 0.003$ &  $\mathbf{ -1.473 \pm 0.003 }$ \\
RMSE & Rt.-Baxter &   &             $0.583 \pm 0.015$ &   $0.609 \pm 0.028$ &   $\mathbf{ 0.106 \pm 0.001 }$ &              $0.152 \pm 0.002$ \\
     & Fw.-Baxter &   &             $0.043 \pm 0.000$ &   $0.053 \pm 0.001$ &   $\mathbf{ 0.035 \pm 0.000 }$ &              $0.036 \pm 0.000$ \\
     & Fw.-Kuka &   &  $\mathbf{ 0.089 \pm 0.000 }$ &   $0.093 \pm 0.000$ &              $0.115 \pm 0.001$ &              $0.115 \pm 0.001$ \\
     & Sarcos &   &             $0.105 \pm 0.000$ &   $0.110 \pm 0.000$ &   $\mathbf{ 0.076 \pm 0.001 }$ &              $0.077 \pm 0.001$ \\
\bottomrule
\end{tabular}

  }
\end{table}

\begin{table}
  \centering
  \caption{
    A compilation of all binary (left) and multi-class (right) classification results from Sections~\ref{sec:binexp} and \ref{sec:multiexp}.
    Numbers are averages $\pm$ standard errors over dataset splits.
  }
  \label{tab:classification_full}
  \resizebox{0.56\linewidth}{!}{%
    \begin{tabular}{ccccccc}
\toprule
      &              &   &                          SVGP &                          PGVI &                       LOO-$k$ \\
Metric & Dataset &           &                               &                               &                               \\
\midrule
\midrule
NLL & Titanic &   &  $\mathbf{ 0.482 \pm 0.019 }$ &  $\mathbf{ 0.483 \pm 0.018 }$ &  $\mathbf{ 0.485 \pm 0.021 }$ \\
      & Bank &   &  $\mathbf{ 0.241 \pm 0.006 }$ &  $\mathbf{ 0.246 \pm 0.007 }$ &  $\mathbf{ 0.234 \pm 0.008 }$ \\
      & Spambase &   &  $\mathbf{ 0.213 \pm 0.004 }$ &  $\mathbf{ 0.217 \pm 0.002 }$ &             $0.222 \pm 0.003$ \\
      & Musk-2 &   &             $0.126 \pm 0.003$ &  $\mathbf{ 0.114 \pm 0.003 }$ &  $\mathbf{ 0.111 \pm 0.005 }$ \\
      & Adult &   &  $\mathbf{ 0.309 \pm 0.002 }$ &             $0.319 \pm 0.002$ &             $0.329 \pm 0.002$ \\
      & Higgs &   &  $\mathbf{ 0.617 \pm 0.002 }$ &  $\mathbf{ 0.619 \pm 0.002 }$ &             $0.629 \pm 0.001$ \\
      & Susy &   &  $\mathbf{ 0.440 \pm 0.001 }$ &  $\mathbf{ 0.442 \pm 0.002 }$ &             $0.468 \pm 0.002$ \\
      & Covtype-Fir &   &             $0.377 \pm 0.001$ &             $0.354 \pm 0.000$ &  $\mathbf{ 0.057 \pm 0.001 }$ \\
      & Covtype-Pine &   &             $0.381 \pm 0.001$ &             $0.357 \pm 0.001$ &  $\mathbf{ 0.060 \pm 0.001 }$ \\
Error & Titanic &   &  $\mathbf{ 0.210 \pm 0.011 }$ &  $\mathbf{ 0.209 \pm 0.011 }$ &  $\mathbf{ 0.210 \pm 0.011 }$ \\
      & Bank &   &  $\mathbf{ 0.098 \pm 0.004 }$ &  $\mathbf{ 0.102 \pm 0.004 }$ &  $\mathbf{ 0.101 \pm 0.006 }$ \\
      & Spambase &   &  $\mathbf{ 0.067 \pm 0.002 }$ &  $\mathbf{ 0.066 \pm 0.002 }$ &             $0.076 \pm 0.003$ \\
      & Musk-2 &   &             $0.036 \pm 0.001$ &  $\mathbf{ 0.033 \pm 0.001 }$ &  $\mathbf{ 0.034 \pm 0.002 }$ \\
      & Adult &   &  $\mathbf{ 0.142 \pm 0.001 }$ &             $0.147 \pm 0.002$ &             $0.146 \pm 0.001$ \\
      & Higgs &   &  $\mathbf{ 0.340 \pm 0.002 }$ &  $\mathbf{ 0.341 \pm 0.002 }$ &             $0.348 \pm 0.001$ \\
      & Susy &   &  $\mathbf{ 0.202 \pm 0.001 }$ &  $\mathbf{ 0.205 \pm 0.001 }$ &             $0.219 \pm 0.001$ \\
      & Covtype-Fir &   &             $0.171 \pm 0.000$ &             $0.161 \pm 0.001$ &  $\mathbf{ 0.023 \pm 0.000 }$ \\
      & Covtype-Pine &   &             $0.174 \pm 0.001$ &             $0.163 \pm 0.001$ &  $\mathbf{ 0.024 \pm 0.000 }$ \\
\bottomrule
\end{tabular}

  }
  \resizebox{0.43\linewidth}{!}{%
    \begin{tabular}{cccccc}
\toprule
      &         &   &                          SVGP &                       LOO-$k$ \\
Metric & Dataset &           &                               &                               \\
\midrule
\midrule
NLL & Cardio &   &  $\mathbf{ 0.387 \pm 0.018 }$ &             $0.466 \pm 0.020$ \\
      & Molec &   &             $0.030 \pm 0.004$ &  $\mathbf{ 0.025 \pm 0.001 }$ \\
      & Abalone &   &  $\mathbf{ 0.691 \pm 0.013 }$ &  $\mathbf{ 0.668 \pm 0.018 }$ \\
      & Wine &   &             $0.996 \pm 0.009$ &  $\mathbf{ 0.945 \pm 0.008 }$ \\
      & Page &   &  $\mathbf{ 0.147 \pm 0.010 }$ &  $\mathbf{ 0.129 \pm 0.010 }$ \\
      & Optical &   &  $\mathbf{ 0.080 \pm 0.005 }$ &             $0.098 \pm 0.004$ \\
      & Landsat &   &  $\mathbf{ 0.197 \pm 0.009 }$ &  $\mathbf{ 0.199 \pm 0.011 }$ \\
      & Thyroid &   &             $0.099 \pm 0.007$ &  $\mathbf{ 0.068 \pm 0.006 }$ \\
      & Pendigits &   &  $\mathbf{ 0.026 \pm 0.004 }$ &  $\mathbf{ 0.032 \pm 0.003 }$ \\
      & Nursery &   &  $\mathbf{ 0.033 \pm 0.001 }$ &  $\mathbf{ 0.033 \pm 0.001 }$ \\
      & Chess &   &  $\mathbf{ 0.034 \pm 0.005 }$ &  $\mathbf{ 0.031 \pm 0.002 }$ \\
      & Shuttle &   &             $0.015 \pm 0.001$ &  $\mathbf{ 0.010 \pm 0.000 }$ \\
      & Covtype &   &             $0.560 \pm 0.002$ &  $\mathbf{ 0.089 \pm 0.001 }$ \\
\midrule
Error & Cardio &   &  $\mathbf{ 0.146 \pm 0.007 }$ &             $0.168 \pm 0.006$ \\
      & Molec &   &  $\mathbf{ 0.001 \pm 0.001 }$ &  $\mathbf{ 0.000 \pm 0.000 }$ \\
      & Abalone &   &             $0.314 \pm 0.008$ &  $\mathbf{ 0.289 \pm 0.008 }$ \\
      & Wine &   &             $0.414 \pm 0.008$ &  $\mathbf{ 0.355 \pm 0.009 }$ \\
      & Page &   &             $0.045 \pm 0.003$ &  $\mathbf{ 0.035 \pm 0.003 }$ \\
      & Optical &   &             $0.020 \pm 0.002$ &  $\mathbf{ 0.014 \pm 0.002 }$ \\
      & Landsat &   &  $\mathbf{ 0.078 \pm 0.005 }$ &  $\mathbf{ 0.075 \pm 0.005 }$ \\
      & Thyroid &   &             $0.044 \pm 0.003$ &  $\mathbf{ 0.019 \pm 0.002 }$ \\
      & Pendigits &   &  $\mathbf{ 0.005 \pm 0.000 }$ &             $0.006 \pm 0.001$ \\
      & Nursery &   &  $\mathbf{ 0.008 \pm 0.001 }$ &             $0.010 \pm 0.001$ \\
      & Chess &   &  $\mathbf{ 0.009 \pm 0.001 }$ &  $\mathbf{ 0.009 \pm 0.001 }$ \\
      & Shuttle &   &             $0.005 \pm 0.001$ &  $\mathbf{ 0.002 \pm 0.000 }$ \\
      & Covtype &   &             $0.238 \pm 0.001$ &  $\mathbf{ 0.032 \pm 0.000 }$ \\
\bottomrule
\end{tabular}

  }
\end{table}

\end{document}